\documentclass[runningheads]{llncs}
\usepackage{graphicx}
\usepackage{amsmath}
\usepackage{array}
\usepackage[caption=false,font=normalsize,labelfont=sf,textfont=sf]{subfig}
\usepackage{url}
\usepackage[pscoord]{eso-pic}
\newcommand{\placetextbox}[3]{
	\setbox0=\hbox{#3}
	\AddToShipoutPictureFG*{
		\put(\LenToUnit{#1\paperwidth},\LenToUnit{#2\paperheight}){\vtop{{\null}\makebox[0pt][c]{#3}}}%
	}%
}%

\begin{document}
\title{Deep Sea Robotic Imaging Simulator}
%
%
\author{Yifan Song\inst{1}
	\and
David Nakath\inst{1} \and
Mengkun She\inst{1} \and
Furkan Elibol\inst{1} \and
Kevin K{\"o}ser\inst{1}}
\authorrunning{Y. Song et al.}
%
\institute{Oceanic Machine Vision,\\ GEOMAR Helmholtz Centre for Ocean Research Kiel, Kiel, Germany\\
\url{https://www.geomar.de/en/omv}\\
\email{\{ysong; dnakath; mshe; felibol; kkoeser\}@geomar.de}}
\maketitle              

\placetextbox{0.5}{0.97}{\fbox{\textsf{This is the pre-print version. The final authenticated version is available online at \url{https://doi.org/10.1007/978-3-030-68790-8_29}}}}%

\begin{abstract}
Nowadays underwater vision systems are being widely applied in ocean research.
However, the largest portion of the ocean - the deep sea -  still remains mostly unexplored.
Only relatively few image sets have been taken from the deep sea due to the physical limitations caused by technical challenges and enormous costs. Deep sea images are very different from the images taken in shallow waters and this area did not get much attention from the community.
The shortage of deep sea images and the corresponding ground truth data for evaluation and training is becoming a bottleneck for the development of underwater computer vision methods. 
Thus, this paper presents a physical model-based image simulation solution, which uses an in-air texture and depth information as inputs, to generate underwater image sequences taken by robots in deep ocean scenarios. 
Different from shallow water conditions, artificial illumination plays a vital role in deep sea image formation as it strongly affects the scene appearance. 
Our radiometric image formation model considers both attenuation and scattering effects with co-moving spotlights in the dark. 
By detailed analysis and evaluation of the underwater image formation model, we propose a 3D lookup table structure in combination with a novel rendering strategy to improve simulation performance. This enables us to integrate an interactive deep sea robotic vision simulation in the Unmanned Underwater Vehicles simulator.
To inspire further deep sea vision research by the community, we release the source code of our deep sea image converter to the public \footnote{\url{https://www.geomar.de/en/omv-research/robotic-imaging-simulator}}.

\end{abstract}
\section{Introduction}
More than 70\% of Earth's surface is covered by water, and more than 90\% of it is deeper than 200 meters, where nearly no natural light reaches. Due to physical obstacles, even nowadays, most of the deep sea is still  unexplored. Deep sea exploration is however receiving increasing attention, as it is the largest living space on Earth, contains interesting resources and is the last uncharted area of our planet.
Since humans cannot easily access this hostile environment, Unmanned Underwater Vehicles (UUVs) have been used for deep sea exploration for decades. With the rapid development of underwater robotic techniques, UUVs are able to autonomously reach and to measure even in several kilometer water depth nowadays, providing platforms for carrying various sensors to explore, measure and map the oceans.

Optical sensors, e.g. cameras, are able to record the seafloor as high resolution images which are advantageous for human interpretation.
Consequently, many UUV platforms are equipped with camera systems for visual mapping of the seafloor due to the significant improvement of imaging capabilities during the last decades.
However, underwater computer vision remains less investigated than on land because underwater images are suffering from several effects, such as attenuation and scattering, which significantly decrease the visibility and the image quality. 
In addition, since no natural light penetrates the deep ocean, artificial light sources are also needed.
This non-homogeneous illumination on limited-size platforms causes anisotropic backscatter that can not be modeled by atmospheric fog models, as often done for shallow water in sunlight, and further degrades image quality.
The above effects often make computer vision solutions struggle or fail in (deep) ocean applications.

The recent trend to employ machine learning methods for various vision tasks even increases the performance gap between underwater vision and approaches on land, since learning methods usually require a large amount of training data to achieve good performance. However, the lack of appropriate underwater (especially deep sea) images with ground truth data is a bottleneck for developing learning-based approaches in this field.
Simulation of deep sea images, in particular with illumination, attenuation and scattering effects could be one way to obtain development or training material for UUV perception.

This paper therefore proposes a physical model-based deep sea underwater image simulator which uses in-air texture images and corresponding depth maps as inputs to simulate synthetic images with underwater optical effects. 
The simulator considers spotlights (with main direction and angular fall-off) and with arbitrary poses in the model for the special conditions in the deep sea.
Several optimization strategies are introduced to improve the computational performance of the simulator, which enables us to integrate the deep sea camera simulation into common underwater robotic simulation platforms (e.g. the Gazebo-based UUV simulator \cite{koenig2004design}).


\section{Related Work and Main Contributions}
\label{Related_Work}

Light rays are attenuated and scattered while traversing underwater volumes, which can be formulated by corresponding radiometric physical models \cite{mobley1994light}.
\cite{jaffe1990computer} and \cite{mcglamery1980computer} decompose underwater image formation into three components: direct signal, forward-scattering and backscatter, which is known as the Jaffe-McGlamery model.
\cite{schechner2004clear} describes the underwater image formation for shallow water cases. Underwater image formation has been intensively studied in underwater image restoration that can be considered as the inverse problem of underwater image formation. 
The most widely applied model has been presented by \cite{cozman1997depth}, which was initially used to recover the depth cues from atmospheric scattering images (e.g. in fog or haze in sunlight):

\begin{equation}
\label{eqSimpleModel}
I = J \cdot e^{-\eta \cdot d}+B \cdot (1-e^{-\eta \cdot d}).
\end{equation}
In the above fog model, the image $I$ is described as a weighted linear combination of object color $J$ and background color $B$. 
Here, $d$ is the distance between the camera and scene point, while $\eta$ represents the attenuation coefficient.

Current underwater image simulators are mostly based on the fog model: \cite{ueda2019underwater} adds a color transmission map and presents a method to generate synthesized underwater images, given an "in-air" image and a depth map that encodes, for each pixel, the distance to the imaged 3D surface. In the literature, such pairs of color images (RGB) and depth maps (D) are also called RGB-D images, and we will use this notation also for the remainder of this paper.
\cite{li2017watergan} proposes a generative adversarial network (GAN) - WaterGAN, which has been trained with shallow water images. It also requires in-air RGB-D images as the input to generate synthetic underwater images. The target function of the GAN discriminator is also based on the fog model.

\begin{figure}
	\centering
	\includegraphics[width=0.3\linewidth]{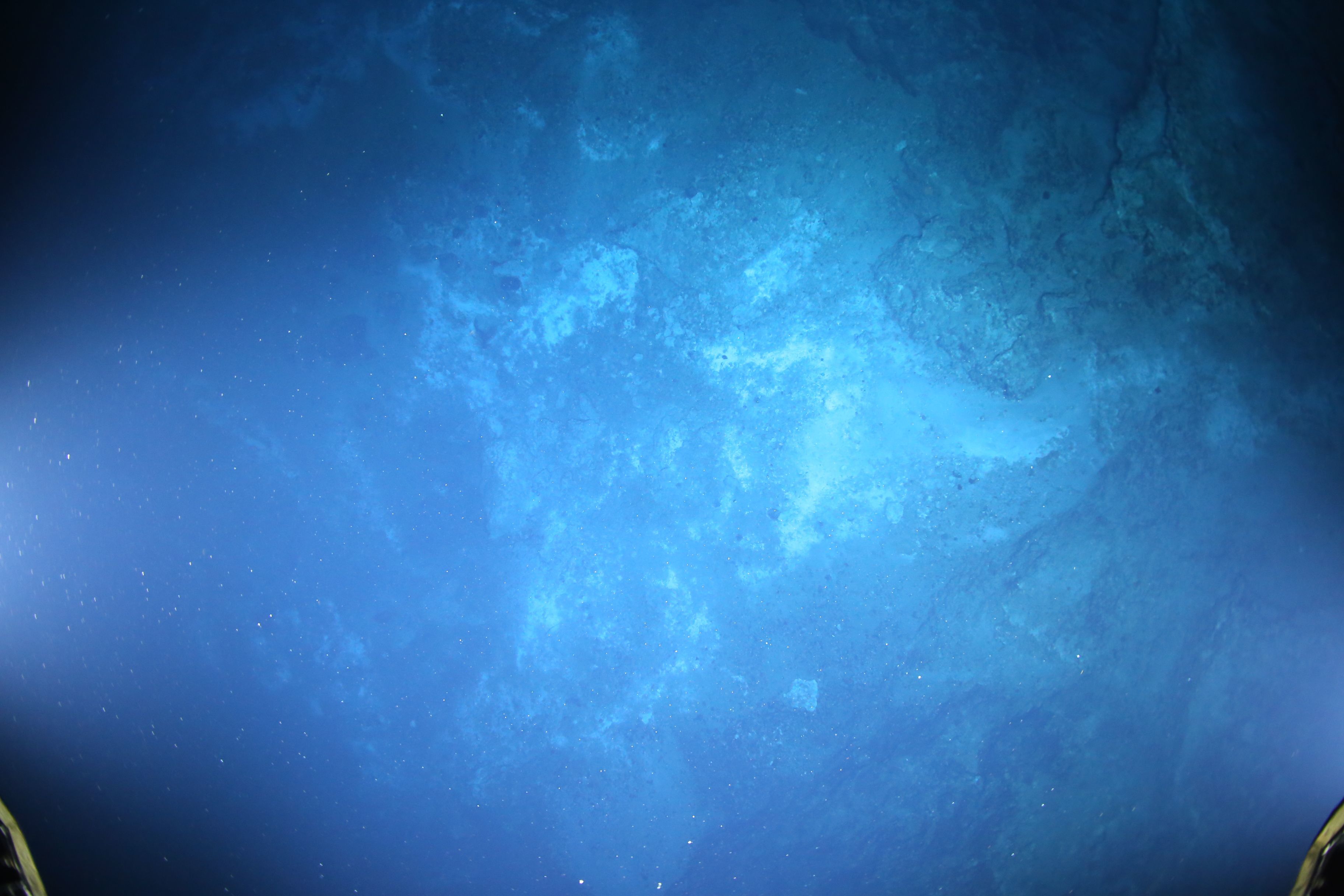}
	\includegraphics[width=0.3\linewidth]{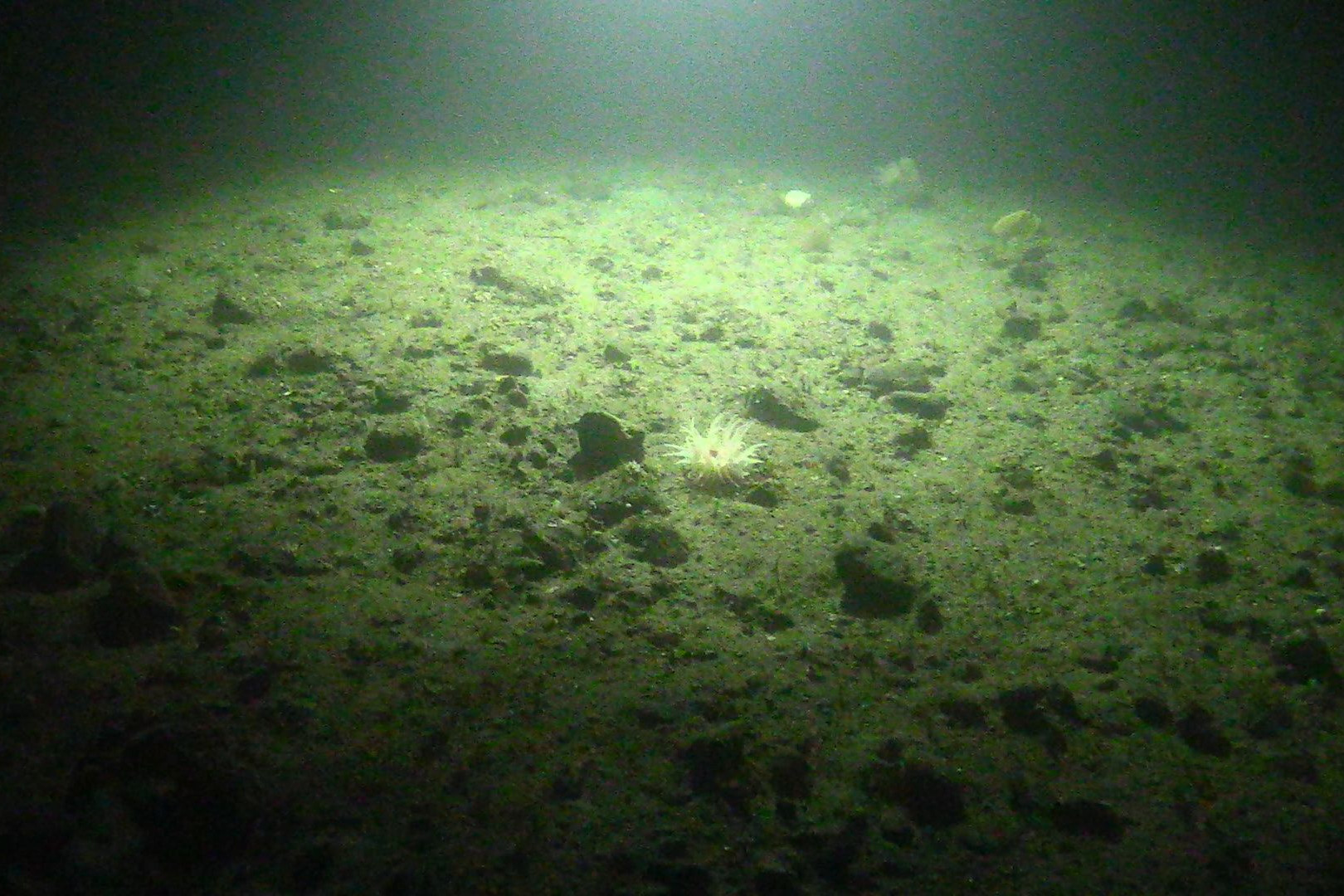}
	\includegraphics[width=0.3\linewidth]{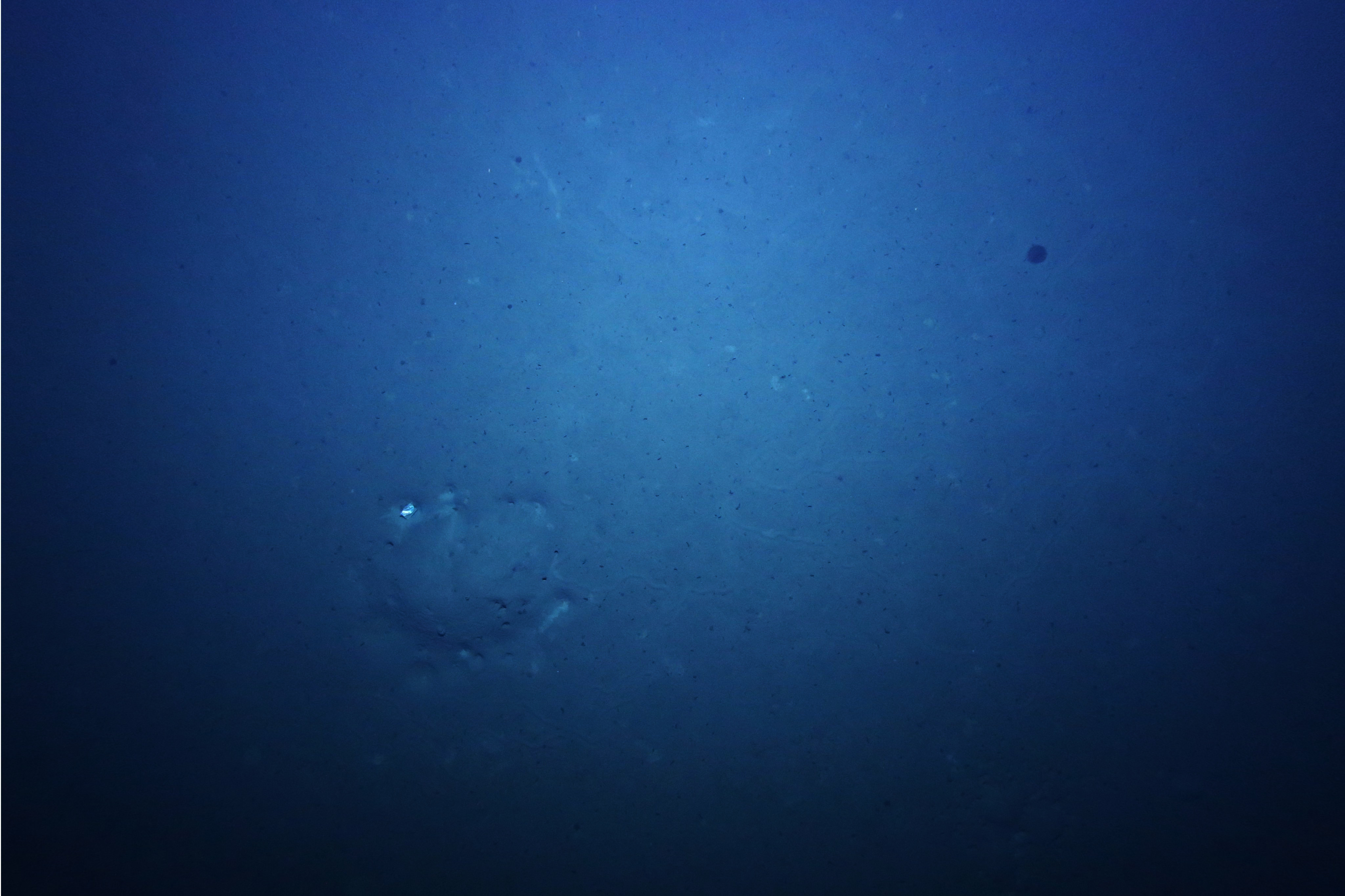}
	\caption{Different artificial lighting configurations strongly affect the appearance of deep sea images, especially the  backscatter pattern (light cones),  which can not be modeled by the fog model. Images courtesy GEOMAR/CSSF/Schmidt Ocean Institute, JAGO Team GEOMAR, AUV Team GEOMAR.}
	\label{fig_lightCone}
\end{figure}

However, the fog model is only valid in shallow water cases, where the scene has global homogeneous illumination from the sunlight. \cite{akkaynak2018revised} addresses many weaknesses of this model, which introduces significant errors in both direct signal and backscatter components.
Obviously, the fog model does not apply to deep sea scenarios where artificial light sources are required to illuminate the scene and the resultant light distribution is extremely inhomogeneous. 
The light originates from the artificial sources attached to the robot and interacts with the water body in front of the camera, leading to very different visual effects in the images, especially in the backscatter component (see Fig. \ref{fig_lightCone}).
Hence, the underwater image formation model in deep sea requires additional knowledge about the light sources like corresponding poses and properties.
\cite{stephan2014computergraphical} uses the recursive rendering equation adapted to underwater imagery considering point light sources in their model.
\cite{sedlazeck2011simulating} proposes an underwater renderer based on physical models for refraction, but not focusing on realistic light sources.
Since backscatter is computed for each pixel for each image, the simulation is quite demanding and does not allow real-time performance.
For image restoration rather than simulation, \cite{bryson2016true} considers a spotlight with Gaussian characteristics in the image formation model and applies it to restore the true color of underwater scenes.
Consequently, there is no simulator available to the community that generates realistic deep sea image sequences at interactive frame rates.

A key use case for deep sea image simulation is integrating it into a UUV simulation platform, 
which enables developing, testing and coordinating performance of underwater robotic systems before risking expensive hardware in real applications. Current ray-tracing solutions are too heavy to integrate to real-time robotic simulation platforms.  
For instance, general robotic simulators provide the simulation of a normal camera and a depth sensor, which can jointly be extended to undewater cases. \cite{prats2012open} developed a software tool called UWSim, for visualization and simulation of underwater robotic missions. This simulator includes a camera system to render the images as seen by underwater vehicles but without any water effect.
\cite{Manhaes_2016} extended the open-source robotics simulator Gazebo to underwater scenarios, called UUV Simulator. This simulator uses so-called RGB-D sensor plugins to generate the depth and color images, and then converts them to underwater scenes by using the fog model (Eq. \ref{eqSimpleModel}). 

Another RGB-D based underwater renderer \cite{alvarez2019generation} applies trained convolutional neural networks to style transfer the image output from \cite{Manhaes_2016} and additionally add forward scattering and haze effect. However, their improvements still rely on the fog model and the haze addition just manually adds two bright spots, which lacks a physical interpretation. \cite{allais2011sofi} integrated the ocean-\-atmosphere radiative transfer (OSOA) model into their simulator SOFI and created look-up tables to compose the back scatter component. However, the OSOA model only describes the sunlight transformation at the ocean-\-atmosphere interface, which is only suitable for shallow water scenarios. 

The main contributions of this paper are: 
(1) A deep sea underwater image simulation solution based on the Jaffe-McGlamery model considering multiple spotlights (with angular characteristics) with corresponding poses and properties.
(2) Analysis of the components in the deep sea image formation model and several optimizations to improve the simulator's performance in particular for rigid robotic configurations.
(3) Integration of the deep sea imaging simulator into the UUV robotic simulator, which can be applied for underwater robotic development and rapid prototyping.
(4) Open source renderer for facilitating the development and testing in underwater vision and robotics communities.

\section{Deep Sea Image Formation Model}
\label{Image_Formation}

\begin{figure}
	\centering
	\includegraphics[width=0.45\linewidth]{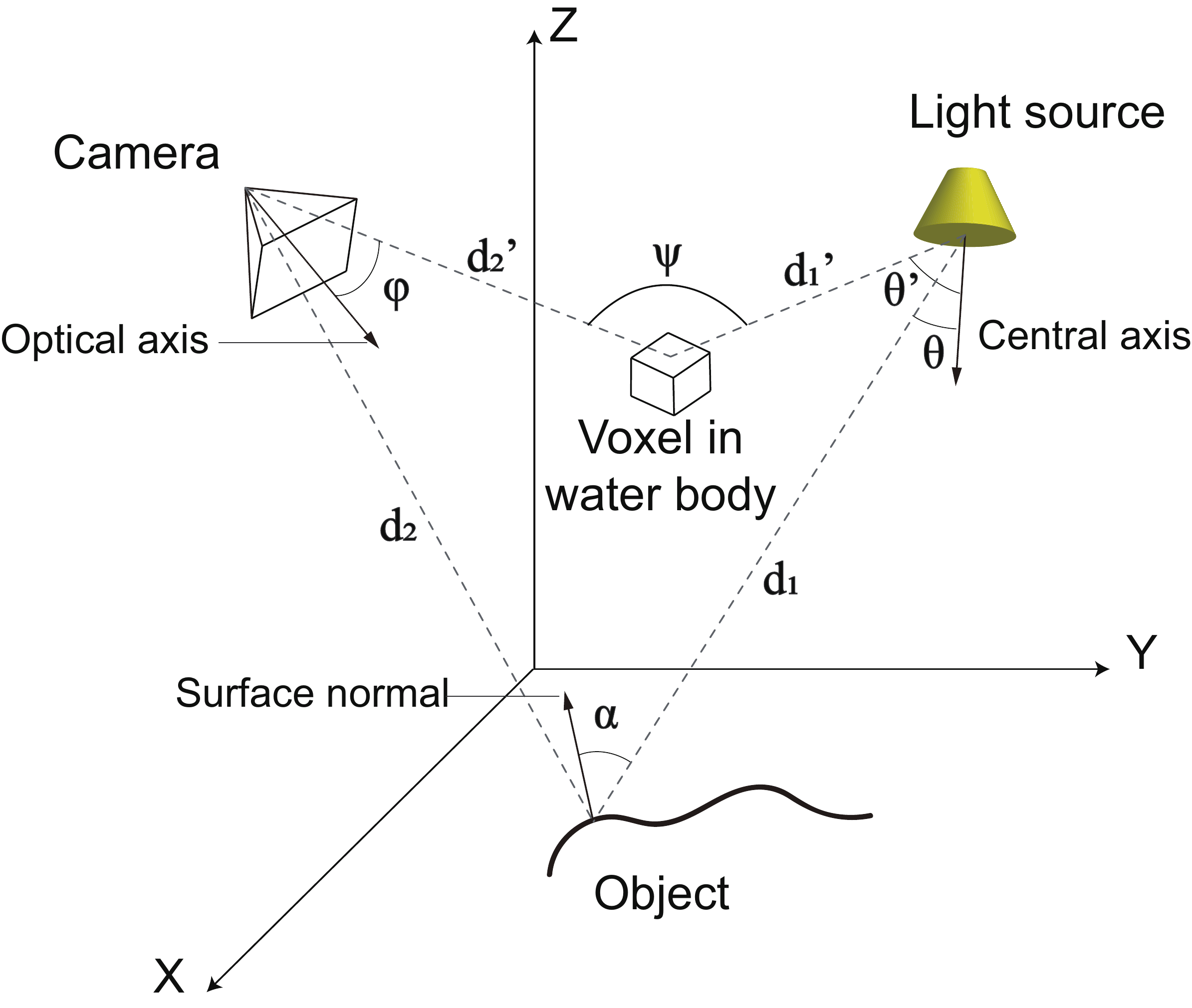}
	\caption{Geometry components involved in the deep sea image formation model (modified from \cite{sedlazeck2011simulating}).}
	\label{fig_model}
\end{figure}

In the deep sea scenario, there is no sun light to illuminate the scene. 
Only artificial light sources, which are attached to the underwater vehicles, provide the illumination.
This moving light source configuration makes the appearance of deep sea images strongly depend on the geometric relationships between the camera, light source and the object (see Fig. \ref{fig_model}). 

\subsection{Radiation of the Light Source}
\begin{figure}
	\centering
	\includegraphics[width=0.55\linewidth]{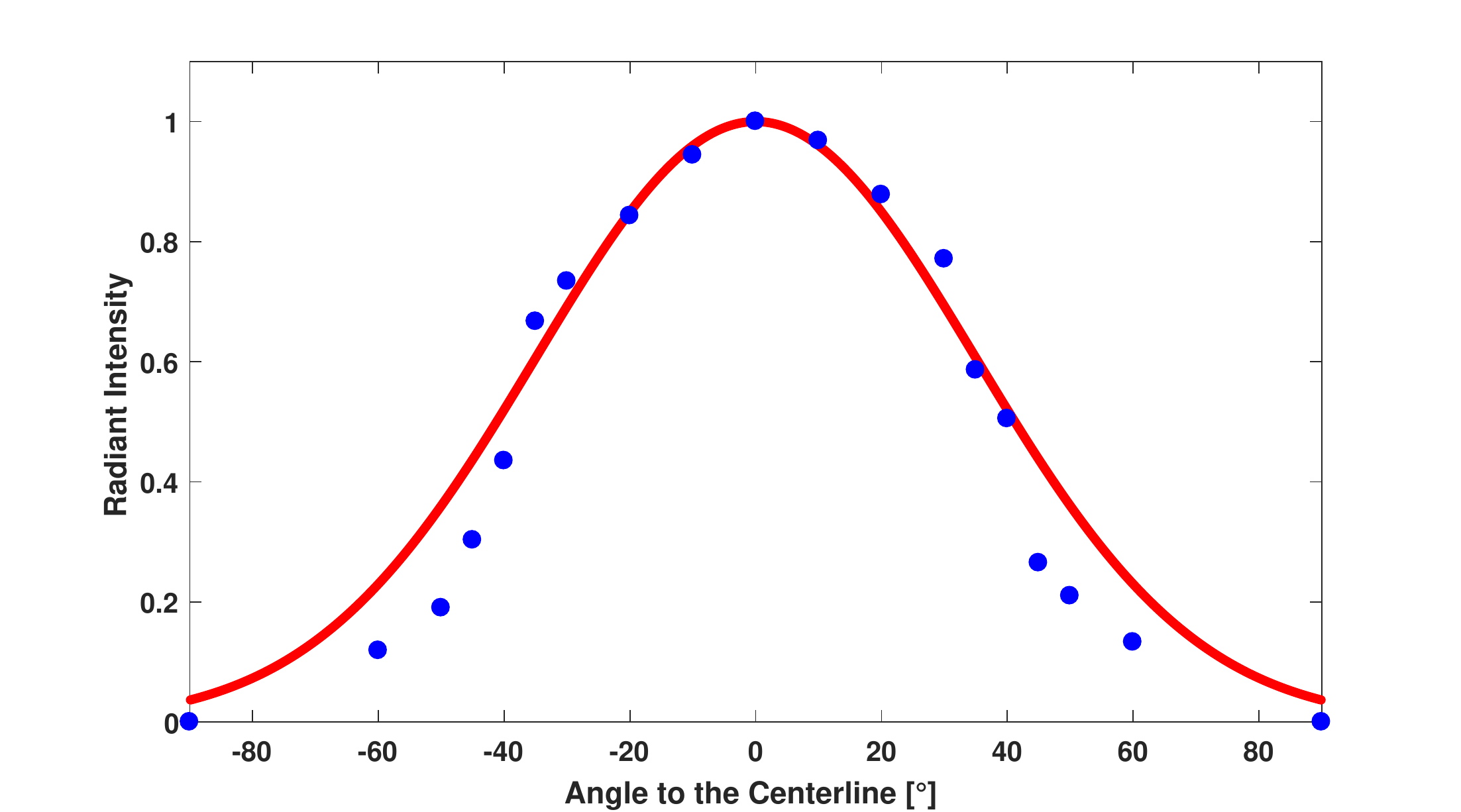}
	\caption{Radiation characteristics of the light source used in this paper, blue dots: our underwater lab measurement, red line: its approximation by using a scaled Gaussian function ($\sigma=35^\circ$).}
	\label{fig_RID}
\end{figure}

This paper considers spotlights, which are commonly used on the UUV platforms. 
This type of light source usually has the highest light emanation along its central axis and an intensity drop-off with increasing angle to the central axis. 
This angular characteristic can be formulated as radiation intensity distribution (RID) curve. 
Often the RID is approximated using a Gaussian function (see e.g. \cite{bryson2016true}). 
In our simulator it is also possible to directly use the sparse measurements as a lookup-table and interpolate the RID values (see Fig. \ref{fig_RID}). 
In the Gaussian model, the radiance along each light ray can be calculated as: 

\begin{equation}
\label{eqLightModel}
I_{\theta}(\lambda) =  I_0(\lambda) e^{-\frac{1}{2} \frac{\theta^{2}}{\sigma^{2}}}.
\end{equation}
Where $I_{\theta}(\lambda)$, $I_0(\lambda)$ are the relative light irradiance at angle $\theta$ and the maximum light irradiance along the central axis respectively. 
The dependency on the wavelength $\lambda$ can be obtained from the color spectrum curve of the LED, which is often provided by the manufacturer or can be measured by a spectrophotometer.

\subsection{Attenuation and Reflection}
\label{Attenuation}
\begin{figure}
	\centering
	\includegraphics[width=0.35\linewidth]{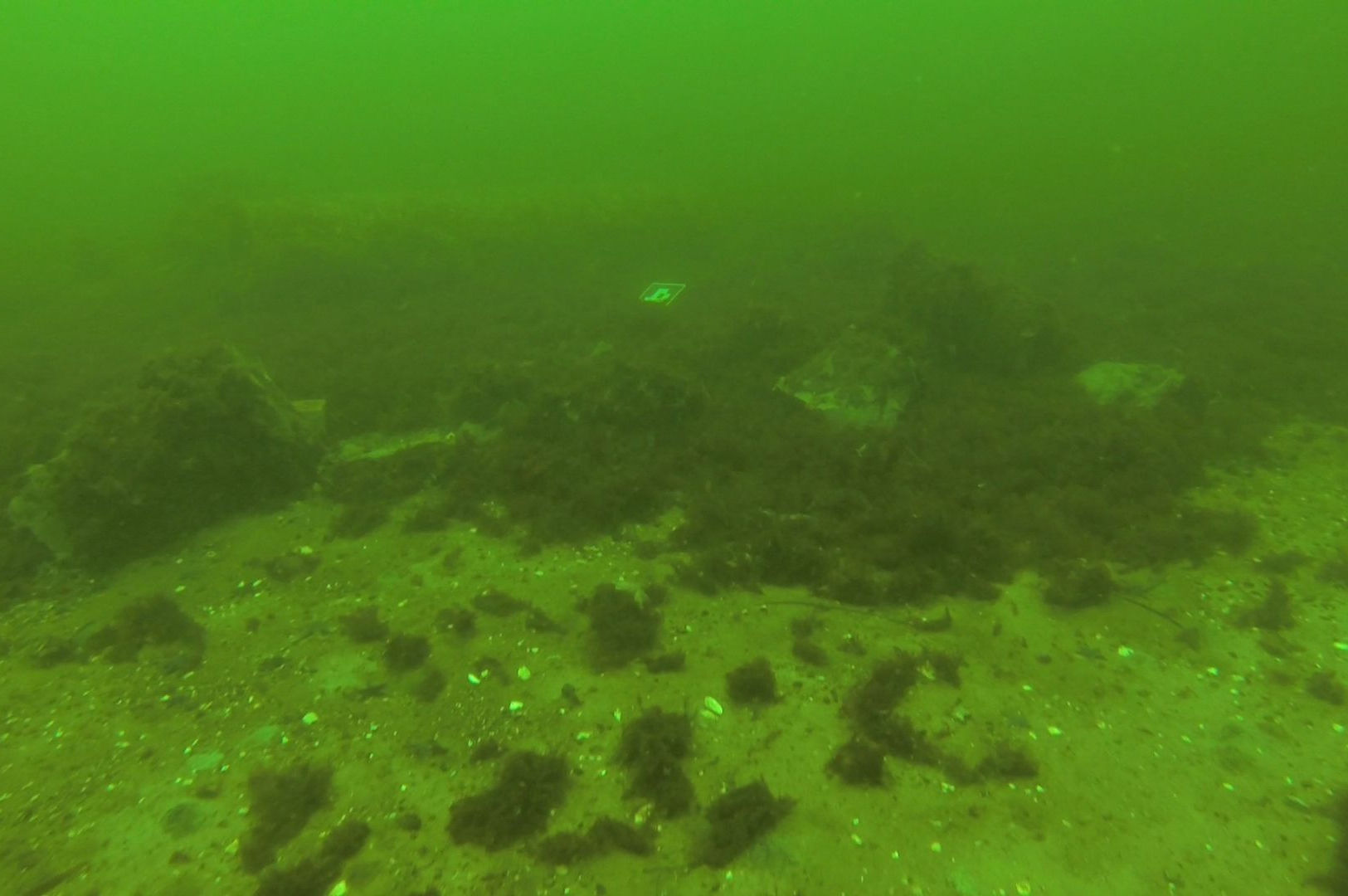} 
	\includegraphics[width=0.35\linewidth]{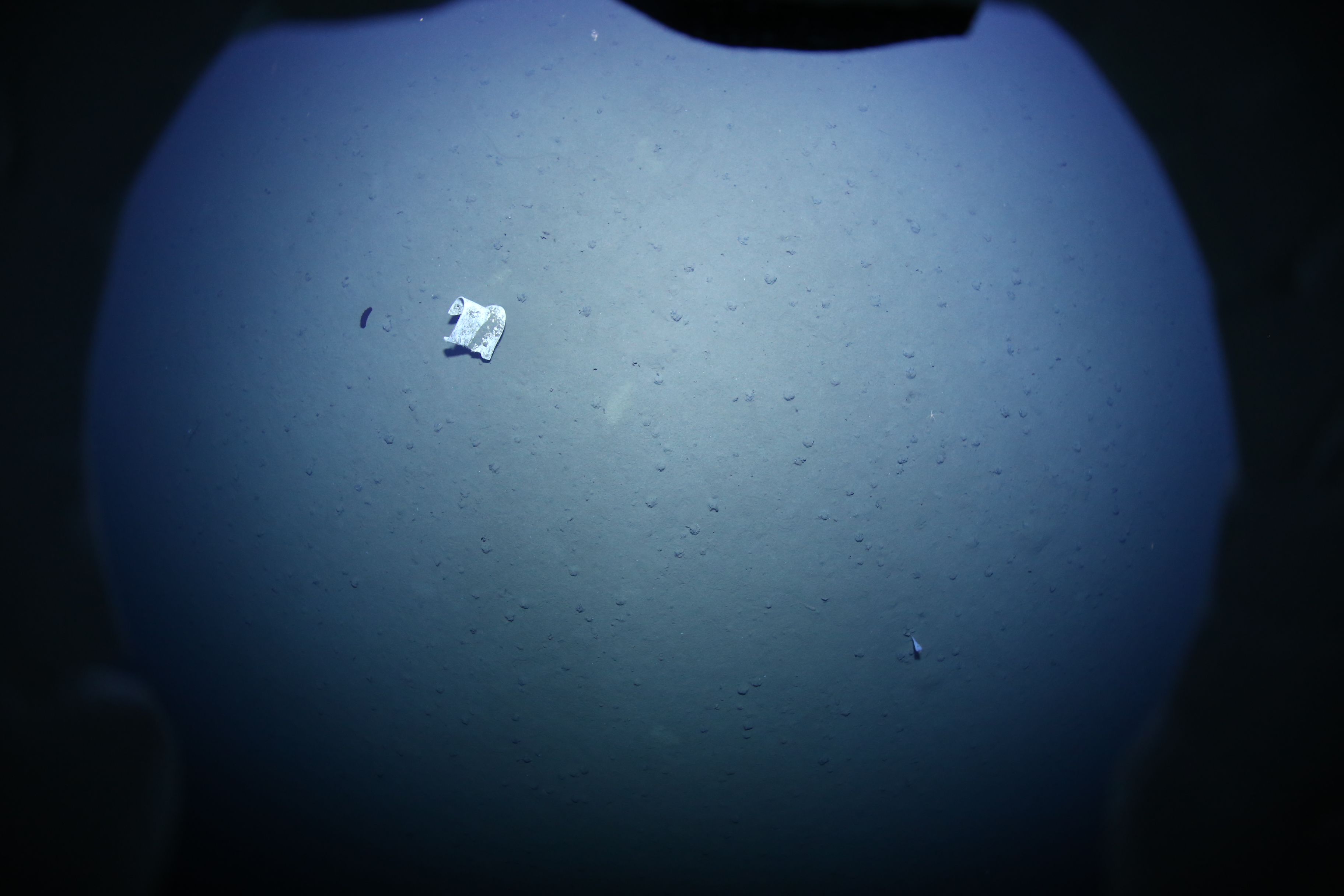}
	
	\caption{Different types of water appear in different colors. Left: coastal water in Baltic Sea. Right: deep sea water in SE Pacific.}
	\label{fig_attenuation}
\end{figure}

Light is attenuated when it travels through the water, where the loss of irradiance depends on the traveling distance and the water properties. 
Different wavelengths of light are absorbed with different strengths, which causes the radiometric changes in underwater images. 
This is because different types of water hold different water attenuation coefficients, resulting in variations of color shifts in images (e.g. coastal water images often appear more greenish, while the deep water images appear more blueish, see Fig. \ref{fig_attenuation}).
\cite{jerlov1968irradiance} measured and classified Earth's waters into five typical oceanic spectra and nine typical coastal spectra. \cite{akkaynak2017space} shows how the corresponding attenuation curves vary between the different types and can serve as a first approximation for typical coefficients (and their expected variations).
Due to the point source property of the spotlight , the Inverse Square Law must be applied in order to simulate the quadratic decay of the light irradiance along the distance from the point-source it originated from.
When we combine the attenuation effect with the object reflection model, which assumes light is reflected equally in all directions on the object surface (Lambertian surface), the entire attenuation and reflection model can be formulated as:

\begin{equation}
\label{eqAttenuation}
E (\lambda) = J(\lambda) \cdot I_{\theta}(\lambda) \frac{ e^{-\eta (\lambda)(d_{1}+d_{2})} } {d_{1}^{2}} \cos{\alpha}.
\end{equation}
Here, $E (\lambda)$ is the irradiance which arrives at the pixel of the image and $J(\lambda)$ is the object color. 
The attenuation parameter $\eta$ indicates the strength of irradiance attenuation through the specific type of water on wavelength $\lambda$. $d_{1}$ and $d_{2}$ refer to the distance from light to object and from object to camera, respectively.
$\alpha$ indicates the incident angle between the light ray from the light source and surface normal.
In the multiple light sources case, the computation is a summation of camera viewing rays for all light sources.
Note that the denominator only contains $d_1$ because with increasing $d_2$ each pixel will simply integrate the light from a larger surface area.

\subsection{Scattering}
\label{Scattering}

The rendering of scattering in this paper is based on the Jaffe-McGlamery model, and is the most complex part of the involved physical models due to its accumulative character.
In the Jaffe-McGlamery model, the scattering is partitioned into two parts: forward scattering and backscatter.
Forward scattering usually describes the light which is scattered by a very small angle, resulting in unsharpness of the scene in the images.
This paper approximates the forward scattering effect with a Gaussian filter $g(\overline{d})$ and the size of filter mask depends on the local scene depth $\overline{d}$. We neglect the forward scattering from light to the scene because the RID curve of the light is usually very smooth (e.g. modeled as a Gaussian function), where a small extra smoothing can be neglected. 
Backscatter refers to light rays which are interacting with ocean water and scattered backwards to the camera, this leads to a "veiling light" effect in the medium.
This effect is happening along the whole light path. 
Following \cite{mcglamery1980computer}, the 3D field in front of the camera can be discretized by slicing it into several slabs with certain thicknesses, the irradiance on each slab is then accumulated in order to form up the backscatter component:

\begin{equation}
\begin{cases}
\label{eqBackScatter}
E^{\prime}(\lambda)= I^{\prime}_{\theta}(\lambda) \frac{ e^{-\eta (\lambda)(d_{1}^{\prime}+d_{2}^{\prime} )} } {d_{1}^{\prime 2}}\\
E_{f}^{\prime}(\lambda)=  E^{\prime}(\lambda) * g(\overline{d_{2}^{\prime}})\\
E_{b}(\lambda )=\sum_{i=1}^{N} \beta(\pi-\psi) [E^{\prime}(\lambda)+E_{f}^{\prime}(\lambda)] \Delta z_{i} \cos(\varphi). \\
\end{cases} 
\end{equation}
Eq. \ref{eqBackScatter} gives the computation of the backscatter component from each light source.
Here $i$ indicates the slab index and $E^{\prime}(\lambda)$ denotes the direct irradiance reaching slab $i$.
$d_{1}^{\prime}$ and $d_{2}^{\prime}$ represent the distances from slab voxel to light source and camera respectively.
$E_{f}^{\prime}(\lambda)$ denotes the forward scattering component of the slab which convolves $E^{\prime}(\lambda)$ by the Gaussian filter $g(\overline{d_{2}^{\prime}})$ and $*$ indicates the convolution operator.
$\beta(\pi-\psi)$ refers to the Volume Scattering Function (VSF), where $\psi$ is the angle between the light ray that hits the voxel and the light ray scattered from the voxel to the camera (see Fig. \ref{fig_model}). The VSF model in this paper applies the measurements from \cite{petzold1972volume} but can be adapted easily to other VSFs. 
$\Delta z_{i}$ is the thickness of the slab and $\varphi$ is the angle between the camera viewing ray and the central axis.

\cite{jaffe1990computer,sedlazeck2011simulating,bryson2016true} also consider optics and electronics of the camera (e.g. vignetting, lens transmittance and sensor response) in their models. 
They are needed to simulate the image of a particular camera and could be added also to our simulator if needed. This is however out of scope for this contribution, where we focus rather on efficient rendering of realistic  backscatter. As discussed in \cite{she2019adjustment}, underwater dome ports can be adjusted in a way to avoid refraction, which is why we also consider adding refraction as a non-mandatory step for underwater simulators (if needed it can be added using the methods proposed in \cite{sedlazeck2011simulating,song2019iterative}).

\section{Implementation}
\label{Implementation}

\begin{figure}
	\centering
	\includegraphics[width=0.48\linewidth]{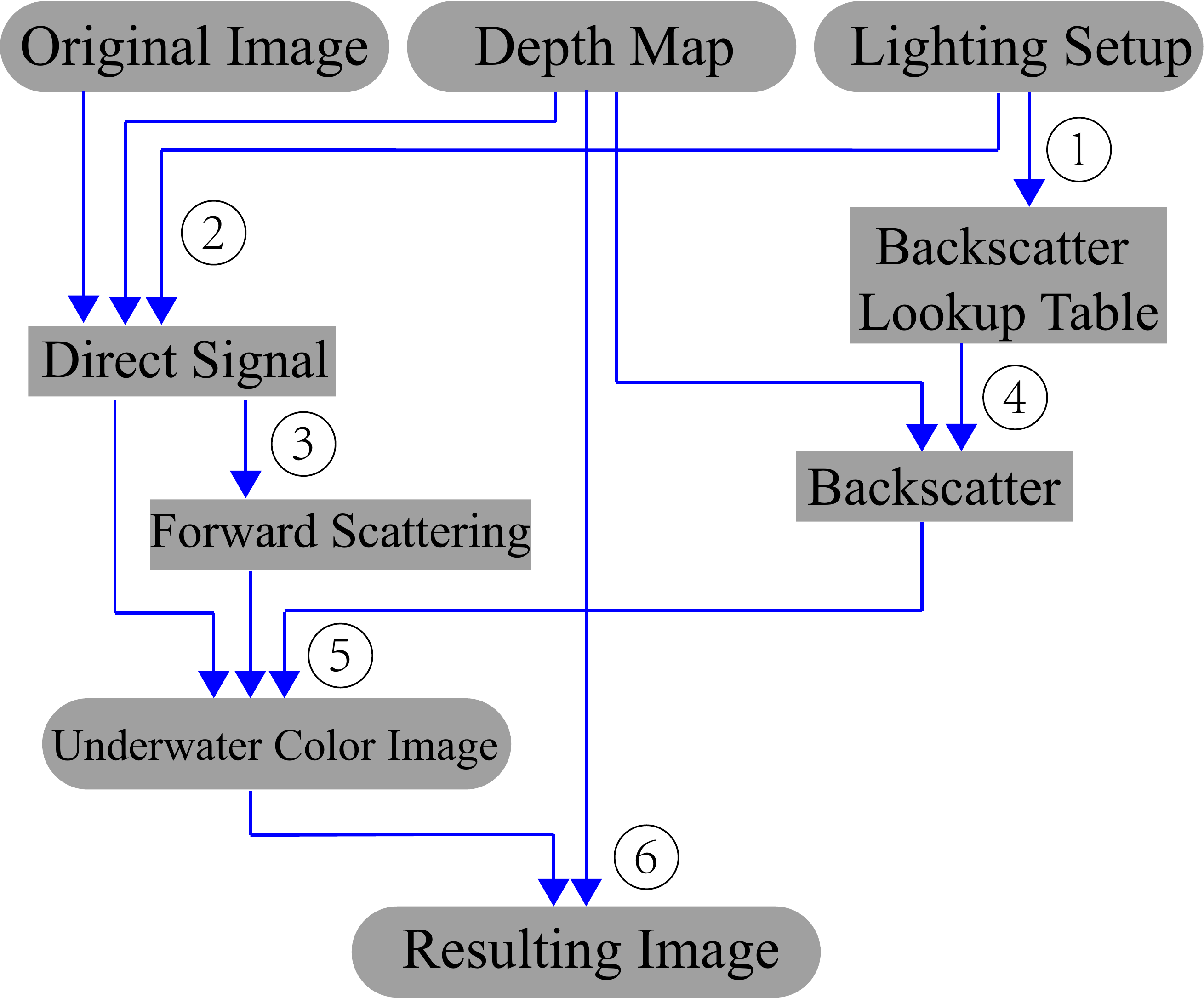}
	\caption{Workflow.}
	\label{fig_workflow}
\end{figure}

This section shows the implementation of our deep sea robotic imaging simulator. The complete workflow is illustrated in Fig. \ref{fig_workflow}. 
\begin{enumerate}
	\item Establish the 3D backscatter lookup table, each unit cell accumulates the backscatter elements along the viewing ray from the camera which is calculated by Eq. \ref{eqBackScatter}.
	\item Generate the direct signal component considering attenuation and object surface reflection according to Eq. \ref{eqAttenuation}.
	\item Compute the forward scattering component by smoothing the direct signal through a Gaussian filter.
	\item Interpolate the backscatter component from the backscatter lookup table with respect to the depth value from the depth map.
	\item Form up the underwater color image by combining the direct signal, forward scattering and the backscatter component.
	\item Optionally, add refraction effect to the image. 
\end{enumerate}
Several optimization procedures are employed in order to improve the performance of the deep sea imaging simulator, as described in the following.

\subsection{Optimizations for Rendering}
In deep sea image simulation, one of the most computationally costly parts is the simulation of the backscatter component. Backscatter happens through the water body between the camera and the 3D scene, which is an accumulative phenomenon in the image. 
However, when the relative geometry between camera and light source is fixed, given the same water, backscatter remains constant in the 3D volume in front of the camera. 
For example, if there are no objects but only water in front of the camera, the image will be relatively constant and only contains the backscatter component.
Once the object appears in the scene, the backscatter volume is cut depending on the depth between the object and the camera, the remaining part is accumulated to form up the image backscatter component.

\begin{figure} [t]
	\centering
	\includegraphics[width=0.5\linewidth]{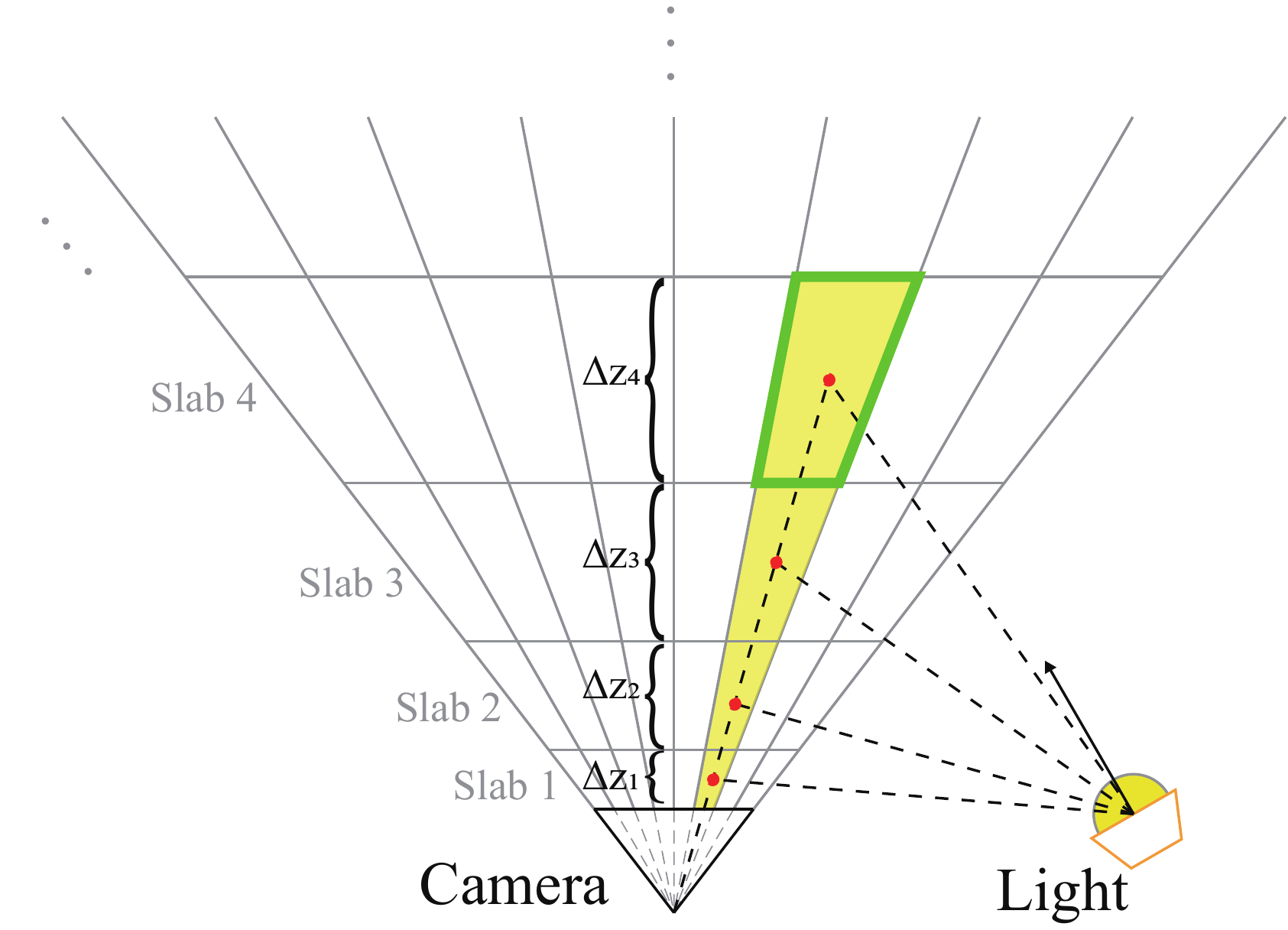}
	\caption{Pre-rendered backscatter field, each unit cell in the slab (green) stores the accumulated backscatter component (yellow) along the camera viewing ray.}
	\label{fig_VoxelField}
\end{figure}
\begin{figure} [t]
	\centering
	\includegraphics[width=0.24\linewidth]{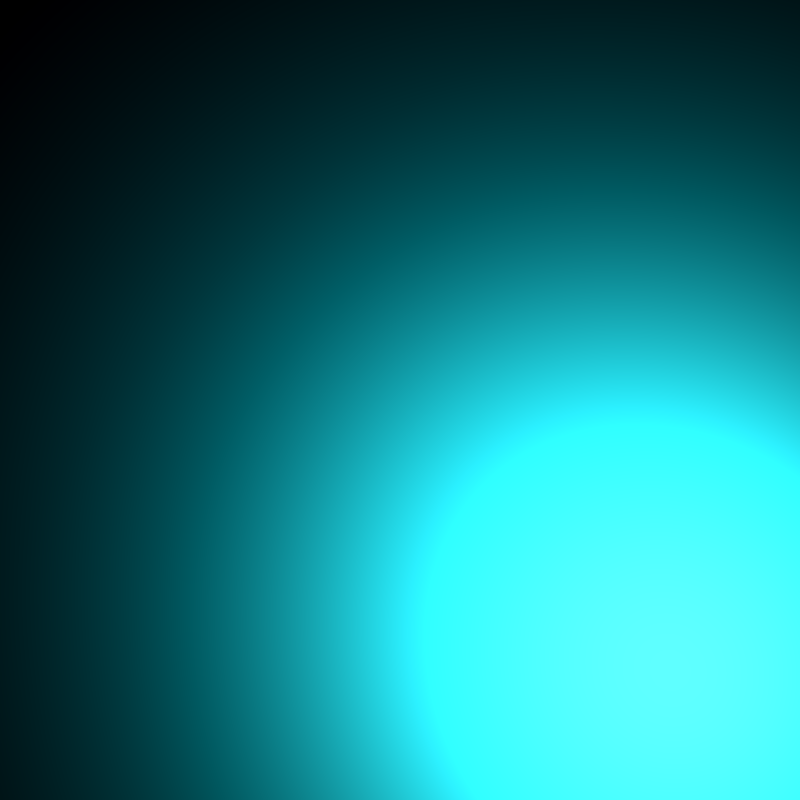}
	\includegraphics[width=0.24\linewidth]{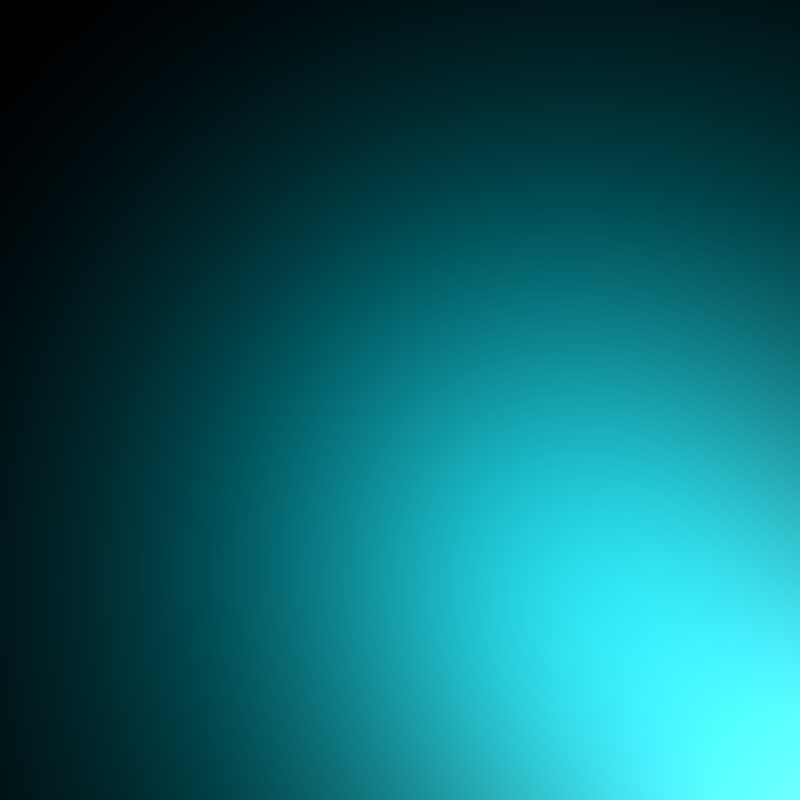}
	\caption{Rendering of backscatter component under the same setups ($d_{max}=10 \rm m$, $N=3$, single light which is at (1m, 1m, 0m) in camera coordinate system and pointing parallel to the camera optical axis.) with different slab thickness sampling approaches. Left: by equal distance sampling, Right: by Eq. \ref{eqSlabSampling}.}
	\label{fig_SlabSampling}
\end{figure}

To this end, we construct a 3D frustum of a pyramid for the camera's field of view and slice it into several volumetric slabs with certain thicknesses parallel to the image plane (see Fig. \ref{fig_VoxelField}). 
Each slab is rasterized into unit cells according to the image size.
We pre-compute the accumulative backscatter elements for each unit cell and store them in a 3D lookup table.
Since the backscatter component of each pixel is an integration of all the illuminated slabs multiplied by the corresponding slab thickness along the viewing ray, the calculation of the backscatter for a pixel with depth $D$ then is simplified by interpolating the value between the closest two unit cells along the viewing ray.

\begin{figure} [t]
	\centering
	\includegraphics[width=0.466\linewidth]{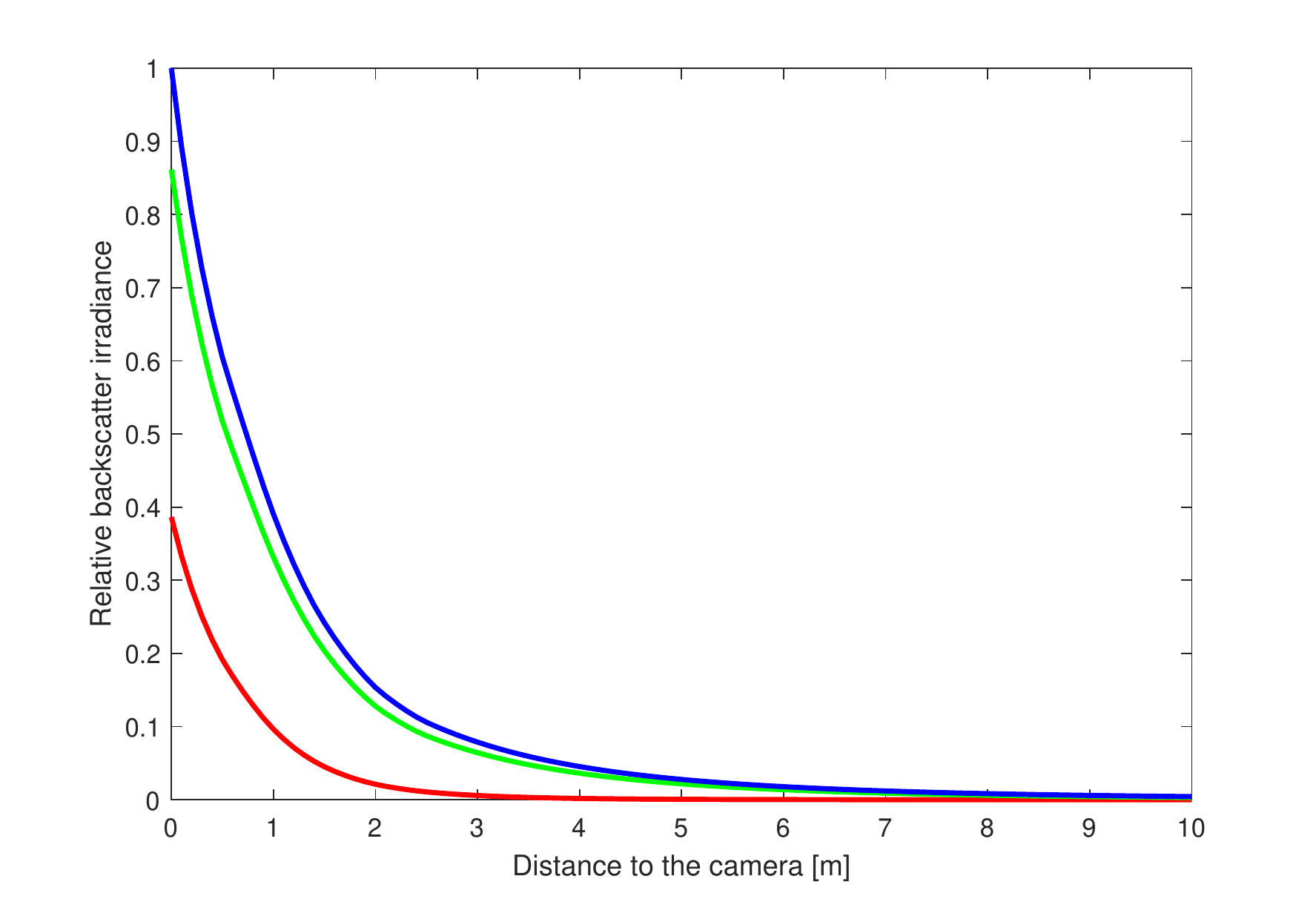}
	\caption{Normalized backscattered irradiance along camera optical axis at different depth of slabs. Each curve describes the backscatter behavior of Jerlov water type II with the same light settings as Fig. \ref{fig_SlabSampling}. It can be seen that in this configuration almost no scattered light reaches the sensor from more than 8m distance. This puts an upper limit on the extent of the lookup table for backscatter.}
	\label{fig_jerlovBackscatter}
\end{figure}
\begin{figure*} [t]
	\begin{center}
		\includegraphics[width=0.2\textwidth]{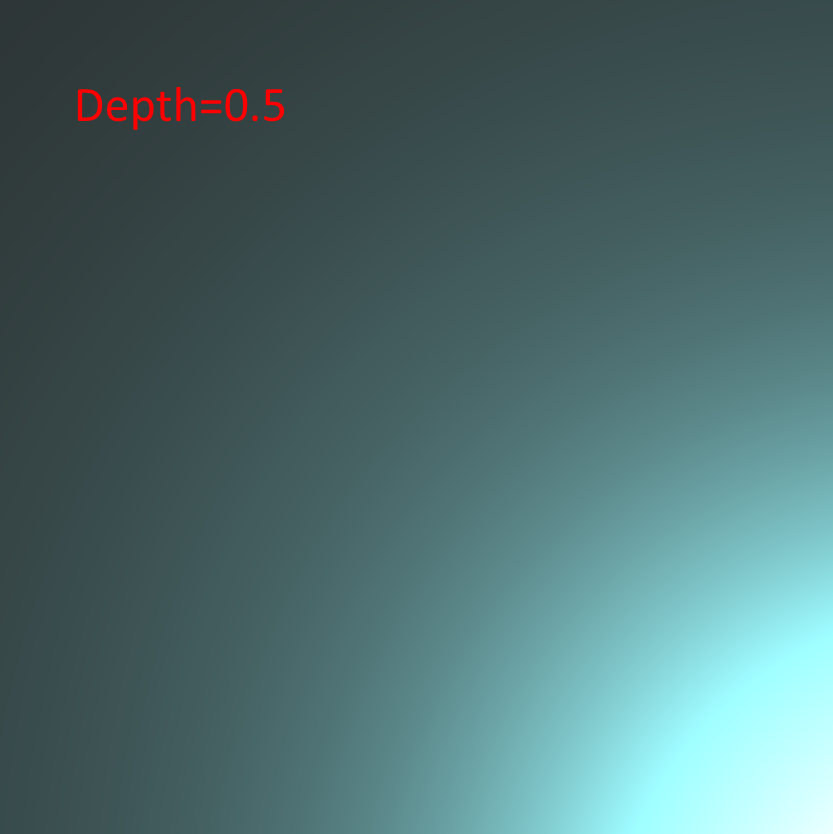} 
		\includegraphics[width=0.2\textwidth]{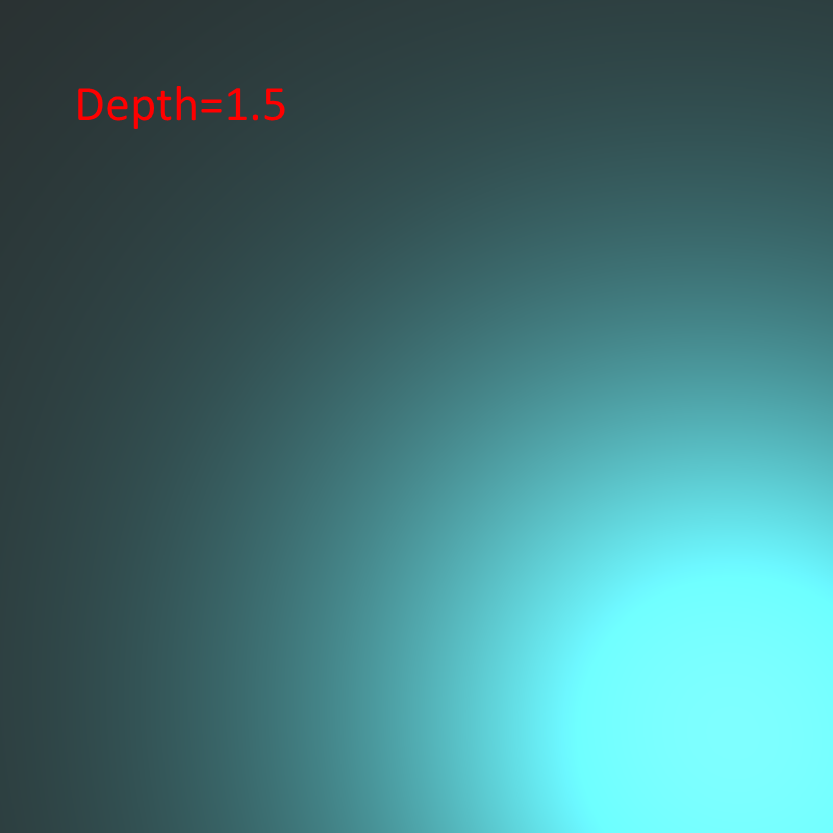} 
		\includegraphics[width=0.2\textwidth]{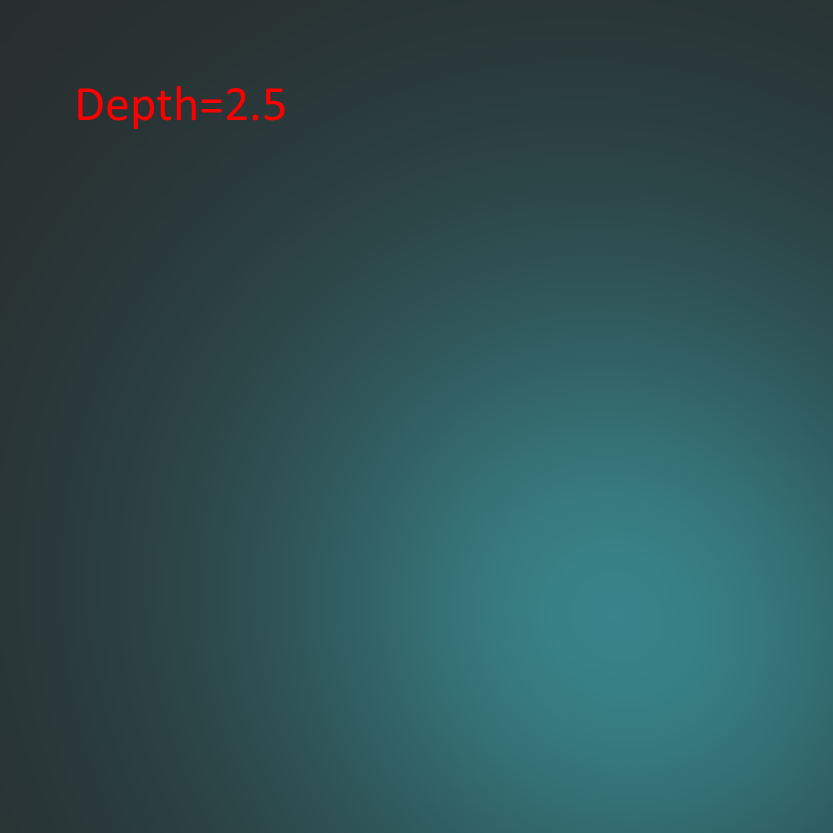}
		\includegraphics[width=0.2\textwidth]{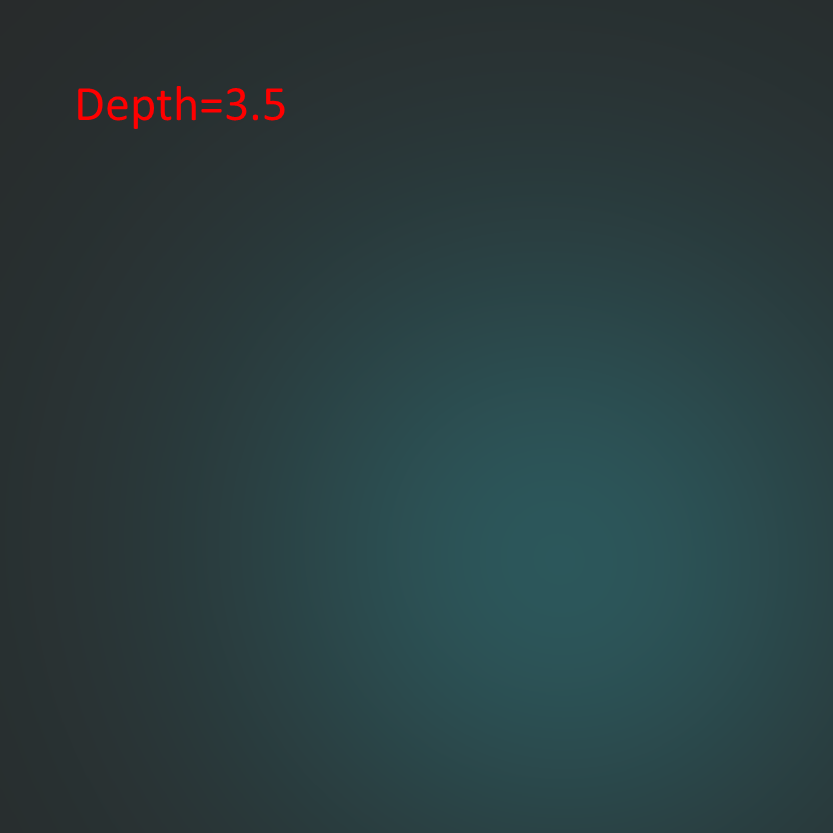}		
		\includegraphics[width=0.2\linewidth]{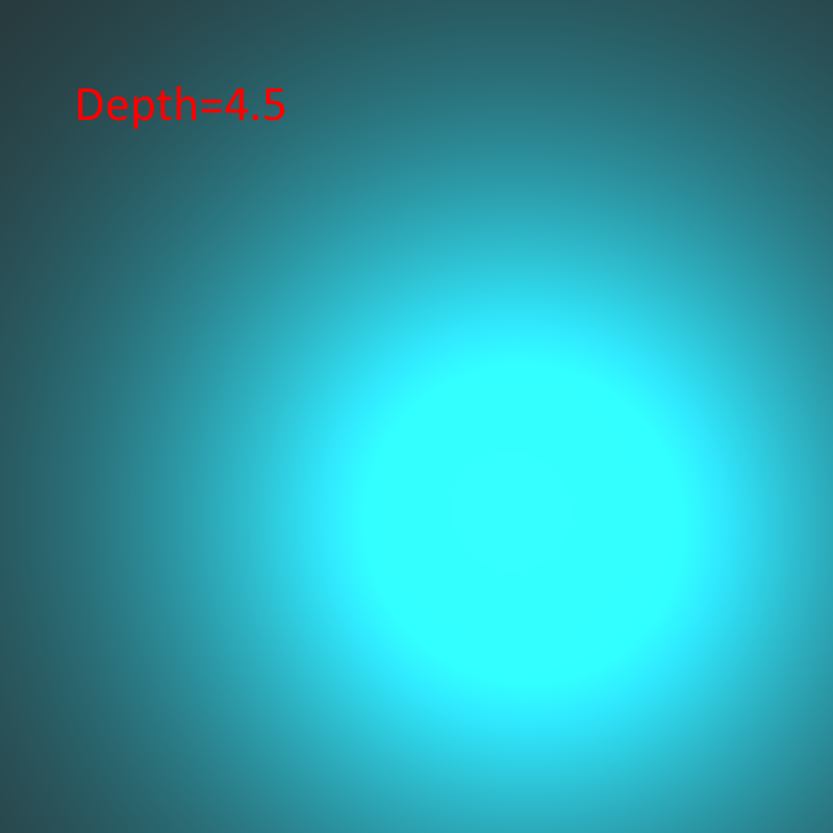}
		\includegraphics[width=0.2\linewidth]{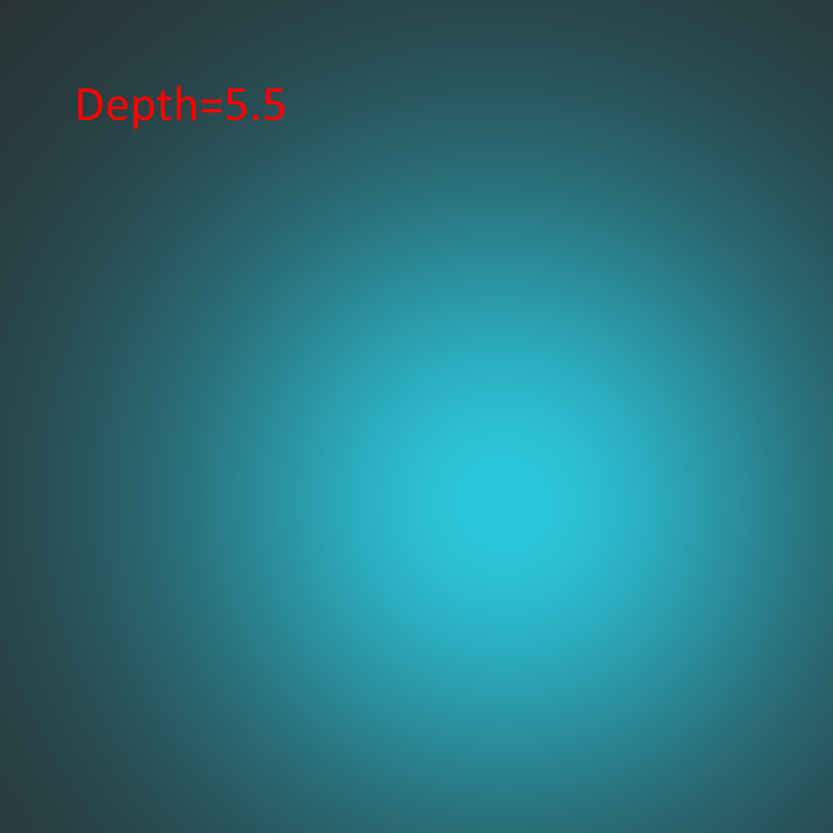}
		\includegraphics[width=0.2\linewidth]{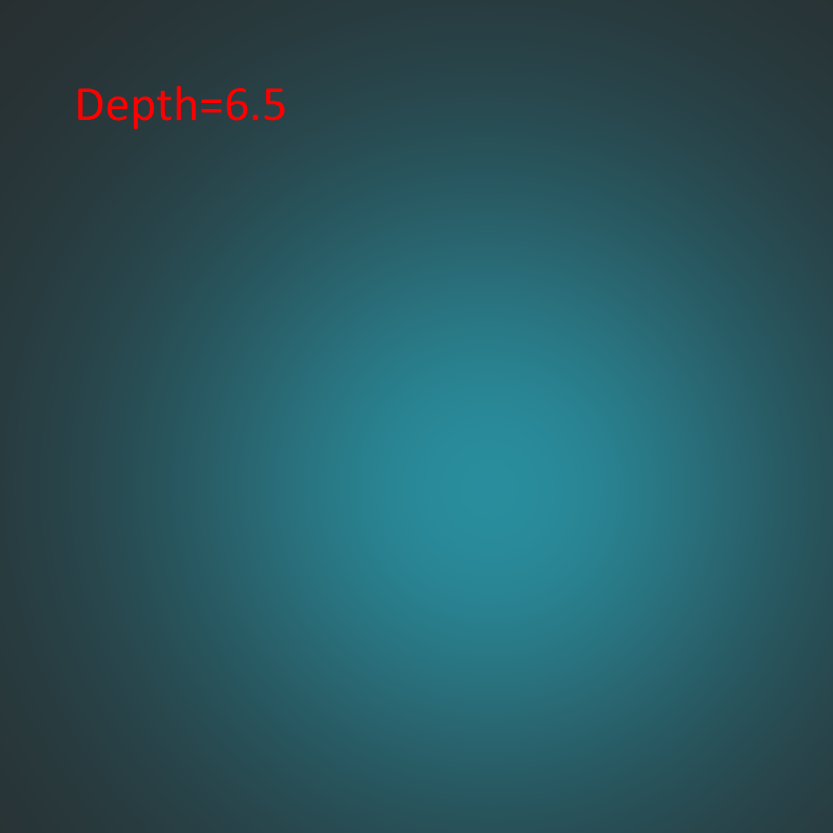}
		\includegraphics[width=0.2\linewidth]{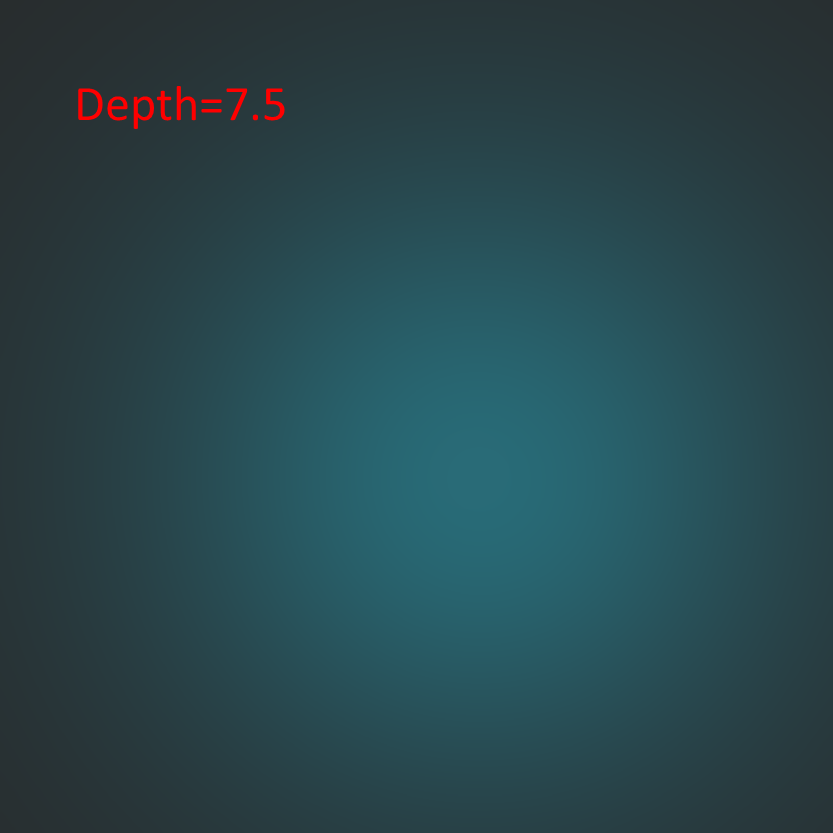}
	\end{center}
	\caption{Backscatter components of different slabs from 0.5m to 7.5m depth (Second row images' intensities are amplified 10 times).}
	\label{fig_slabs}
\end{figure*}

During the rendering of the slabs, we noticed that in practically relevant UUV camera-light configurations, the backscatter component appearance is dominated by the irradiance from the water volume close to the camera and scattering becomes smoother and eventually disappears in the far field. This depends strongly on the relative pose of the light source(s) and is different in each individual camera system but this is a fundamental difference to the shallow water cases, where also far away from the camera a lot of light from the sun is still available. Sample "scatter irradiance" patterns on slabs can be seen in Fig. \ref{fig_slabs}. In order to generate an accurate backscatter component with less number of slabs, we propose an adaptive slab thickness sampling function based on Taylor series expansion of the exponential function:

\begin{equation}
\label{eqSlabSampling}
\Delta z_{i} = s \cdot \frac{N^{(i-1)}}{(i-1)!}  \;\;\;\;\;\;\;\; (i=1,2,...,N)
\end{equation}
where $\Delta z_{i}$ indicates the slab thickness of slab index $i$.
The scale factor $s=2.2 \cdot d_{max}/{e^{N}}$, where $d_{max}$ refers to the maximum depth of the scene field which is divided into number of slabs $N$.
Here, ${e^{N}}$ normalizes the Taylor series and $2.2 \cdot d_{max}$ ensures the slab thickness is monotonically increasing in $(1<i<N)$ and $\sum_{i=1}^{N} \Delta z_{i} \approx d_{max}, (N>3)$.
This equation leads to denser slab samplings closer to the camera. 
As it is shown in Fig. \ref{fig_SlabSampling}, under the light setup described in its caption, the brightest spot should be at the bottom right corner of the image. The sampling of slab thickness by Eq. \ref{eqSlabSampling} gives a more plausible backscatter rendering result than the equal distance sampling approach.

The value of maximum depth of the scene $d_{max}$ is also an important factor which affects the backscatter rendering quality and performance. In Fig. \ref{fig_jerlovBackscatter} we demonstrate the normalized backscattered irradiance of the voxels along the optical center axis in deep ocean water. This figure can be a good reference for finding $d_{max}$ to simulate the underwater images under different conditions or settings.


\begin{figure} [t]
	\centering
	\subfloat[in-air]{\includegraphics[width=0.24\linewidth]{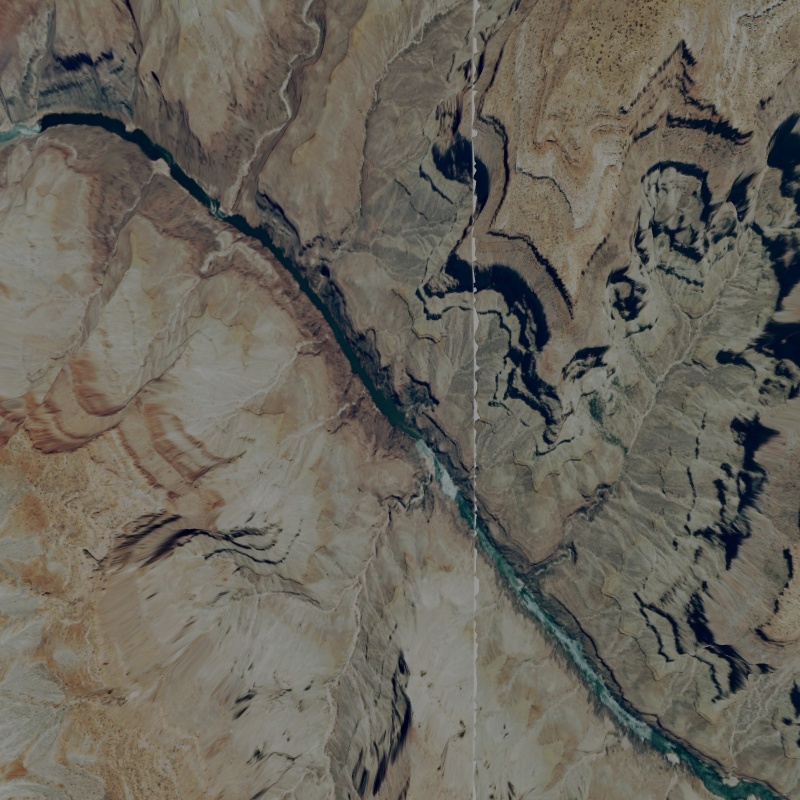}%
		\label{001}}
	\hfil
	\subfloat[depth]{\includegraphics[width=0.24\linewidth]{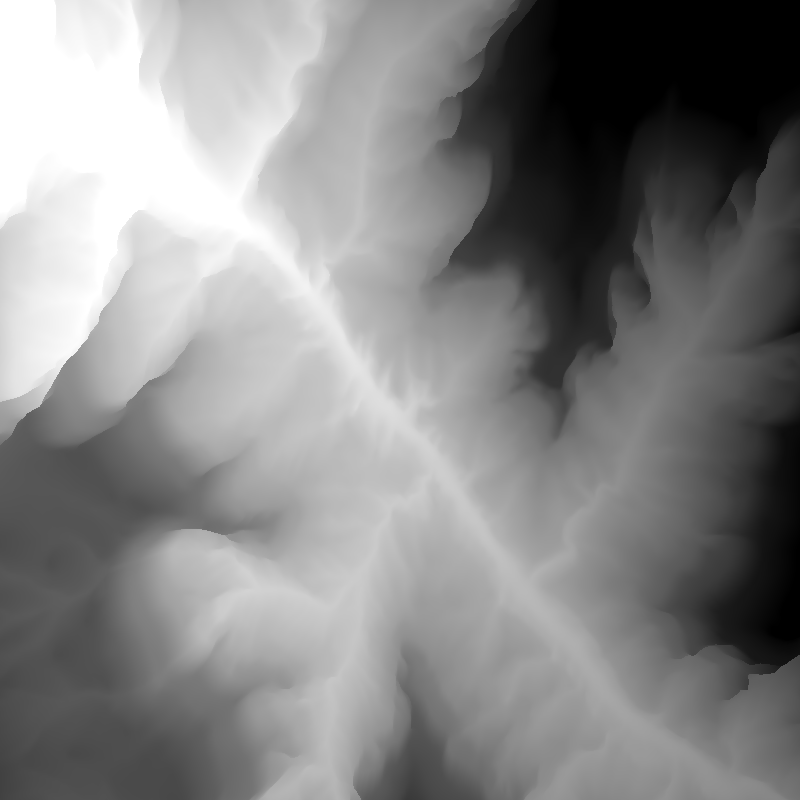}%
		\label{002}} 
	\hfil
	\subfloat[direct signal]{\includegraphics[width=0.24\linewidth]{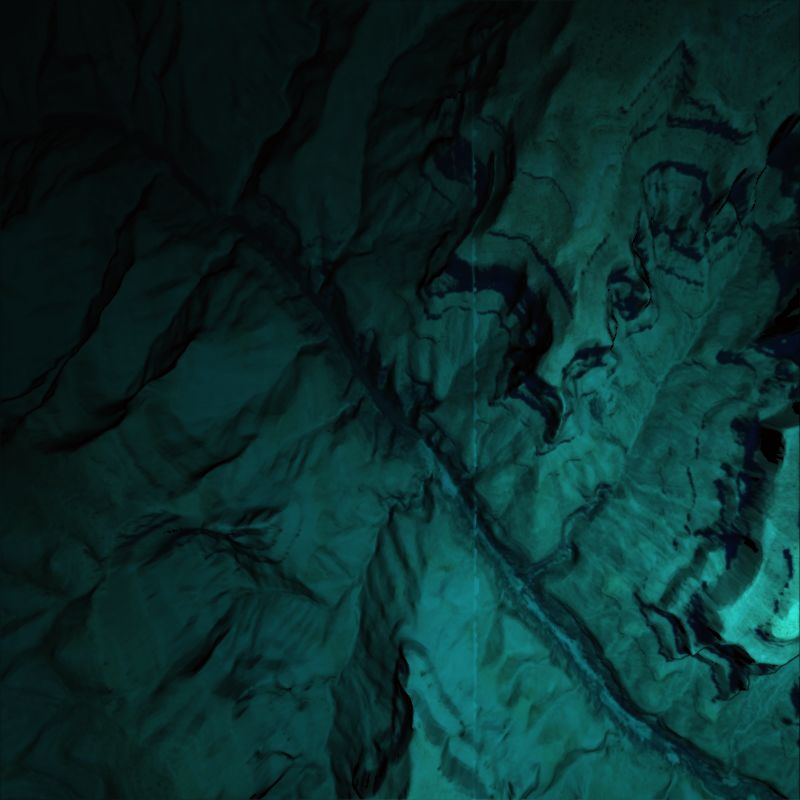}%
		\label{003}}
	\hfil
	\subfloat[backscatter]{\includegraphics[width=0.24\linewidth]{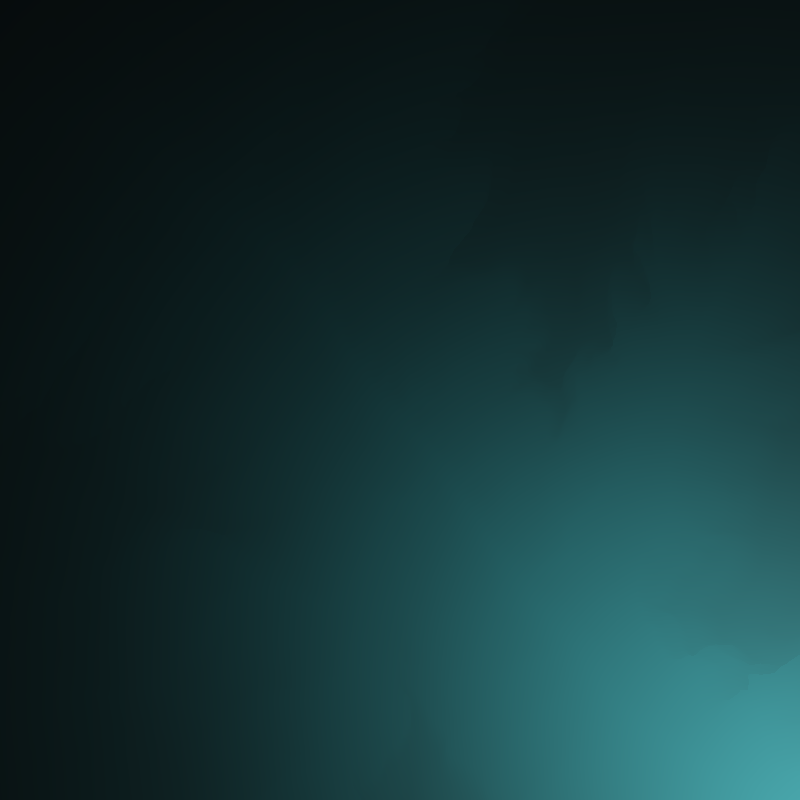}%
		\label{004}} 
	\hfil  \\
	\subfloat[underwater color]{\includegraphics[width=0.24\linewidth]{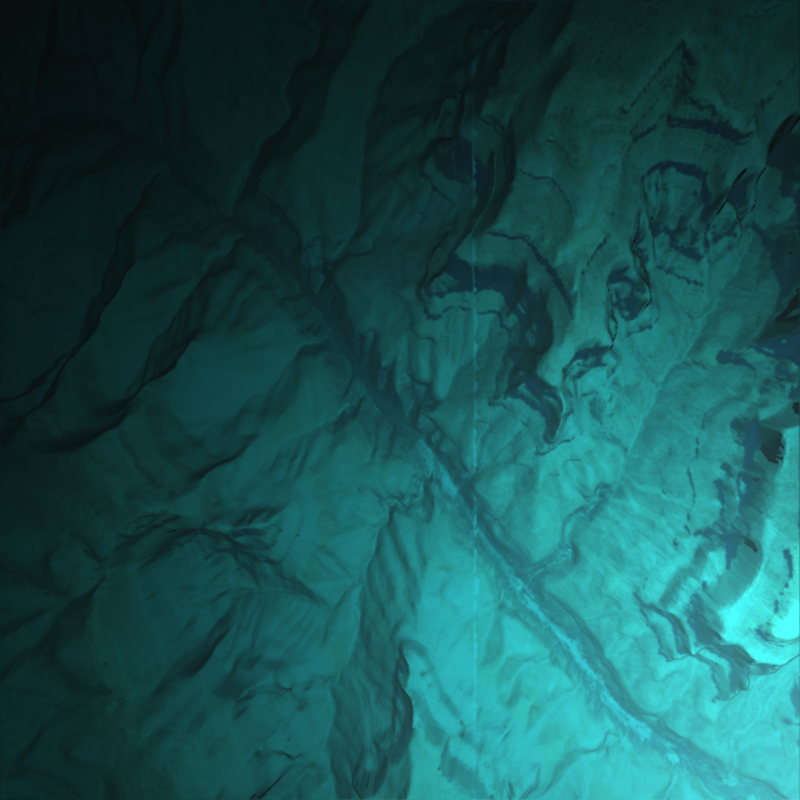}%
		\label{005}}
	\hfil
	\subfloat[add refraction]{\includegraphics[width=0.24\linewidth]{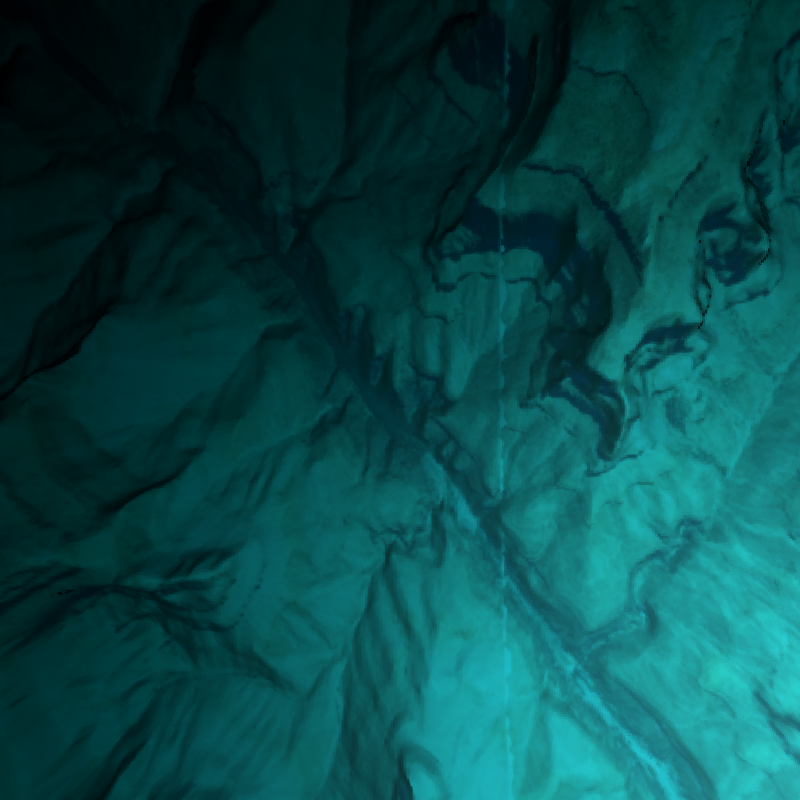}%
		\label{006}}
	\hfil
	\caption{Deep sea image simulation results.}
	\label{fig_rendering}
\end{figure}

\subsection{Rendering Results}
As it is shown in Fig. \ref{fig_rendering}, (a) and (b) are the inputs from the RGB-D sensor plugin. The direct signal (c) and backscatter (d) components are computed respectively, then the simulated underwater color image (e) is constructed by the direct signal, the smoothed direct signal (forward scattering) and the backscatter. In the end, the refraction effect is added to the underwater color image in (f) by using the method from \cite{song2019iterative}.

\begin{figure}[t]
	\begin{center}
		\includegraphics[width=0.95\textwidth]{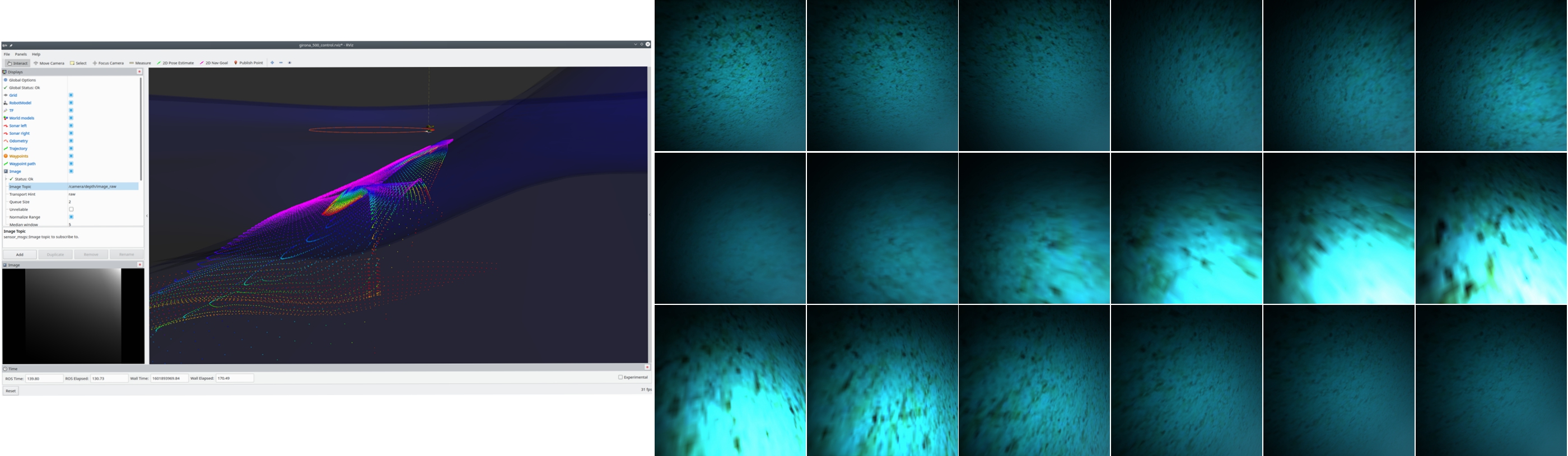} 
	\end{center}
	\caption{Left: camera path overview in simulator. Right: Rendered image sequence. Due to the physically correct model, already in the simulation we can see that some images will be overexposed with the settings chosen. Consequently, the exposure control algorithm of the robot can be adapted already after simulation without wasting precious mission time at sea.}
	\label{fig_gazebo}
\end{figure}

\subsection{Integration in Robotic UUV Simulation Platform}

Gazebo is an open-source robotics simulator. It utilizes one out of four different physics engines to simulate the mechanisms and dynamics of robots. 
Additionally, it provides the platform for hosting various sensor plugins.
\cite{Manhaes_2016} proposes the UUV Simulator which is based on Gazebo and extends Gazebo to underwater scenarios.
The UUV Simulator additionally takes into account the hydrodynamic and hydrostatic forces and moments for simulating vehicle dynamics in underwater environments. 
Several sensor plugins which are commonly deployed on UUVs are also available, including inter alia: inertial measurement unit (IMU), magnetometer, sonar, multi-beam echo sounders and camera modules.
We integrate our deep sea camera simulator into the UUV Simulator camera plugin which provides in-air and depth images as the input and it is able to reach interactive speeds for 800$\times$800 size of images using OpenMP without any GPU acceleration on a 16-core CPU consumer hardware. 
The workspace interface and sample rendering results are shown in Fig. \ref{fig_gazebo}.

\section{Evaluation}
\label{Evaluation}

\begin{figure*}[t]
	\begin{center}
		\subfloat[in-air]{\includegraphics[width=0.3\textwidth]{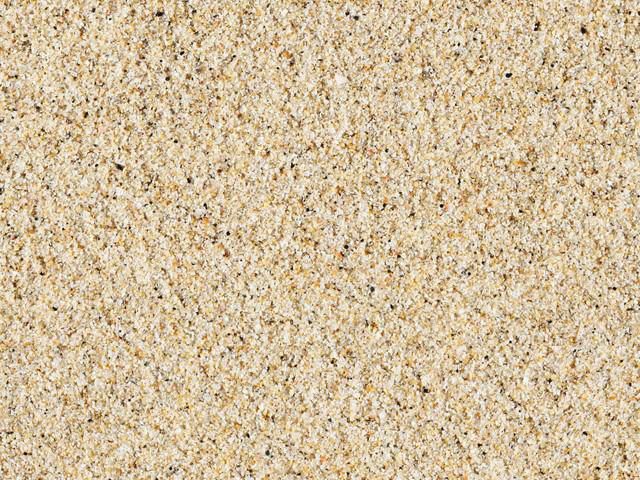}%
			\label{0001}}
		\hfil
		\subfloat[depth]{\includegraphics[width=0.3\textwidth]{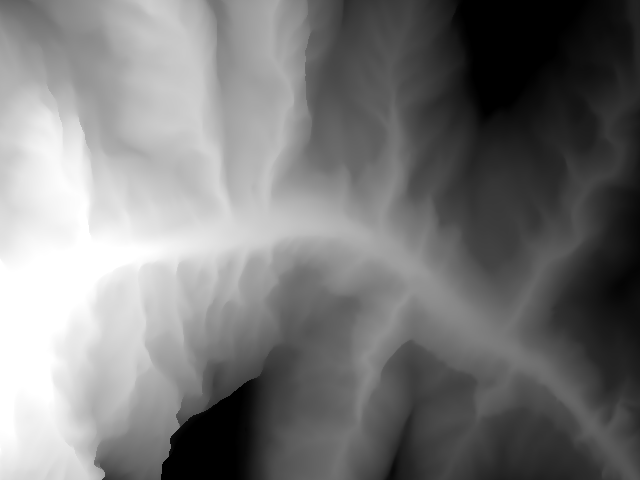}%
			\label{0002}}
		\hfil
		\subfloat[our output]{\includegraphics[width=0.3\textwidth]{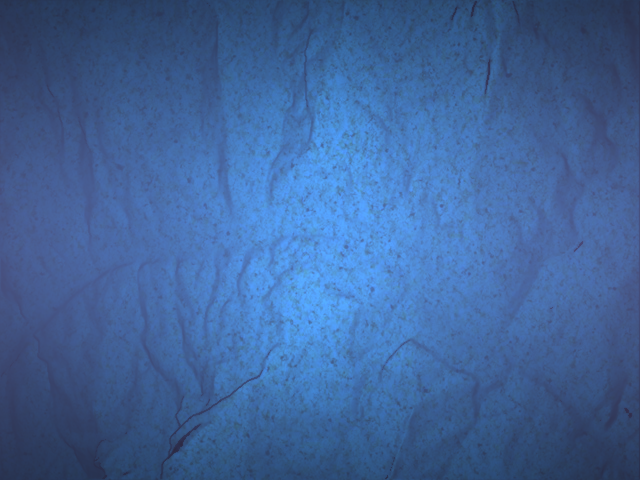}%
			\label{0003}} \\
		\hfil
		\subfloat[in-air shading]{\includegraphics[width=0.24\textwidth]{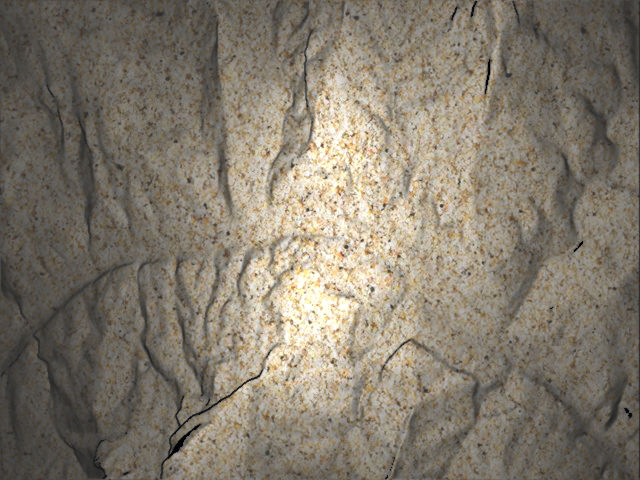}%
			\label{0004}}
		\hfil
		\subfloat[UUV]{\includegraphics[width=0.24\textwidth]{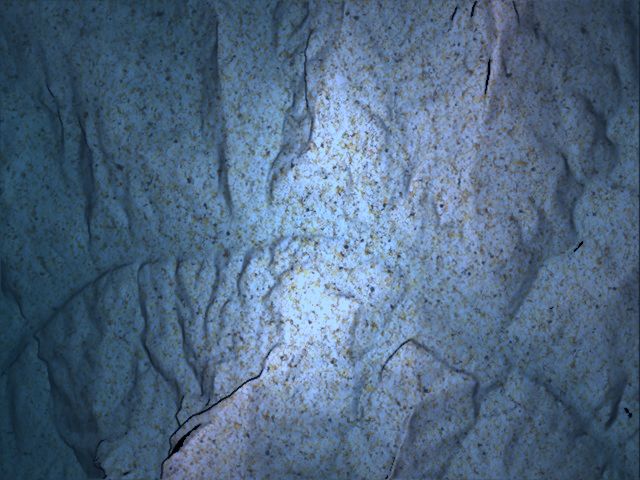}%
			\label{0005}}
		\hfil
		\subfloat[WaterGAN]{\includegraphics[width=0.24\textwidth]{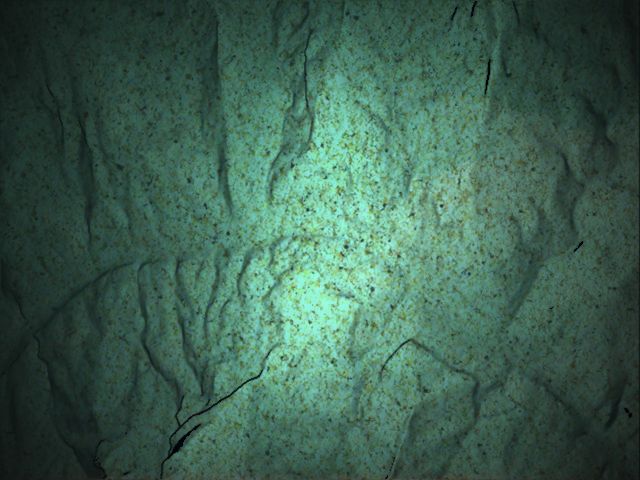}%
			\label{0006}}
		\hfil
		\subfloat[UW\_IMG\_SIM]{\includegraphics[width=0.24\textwidth]{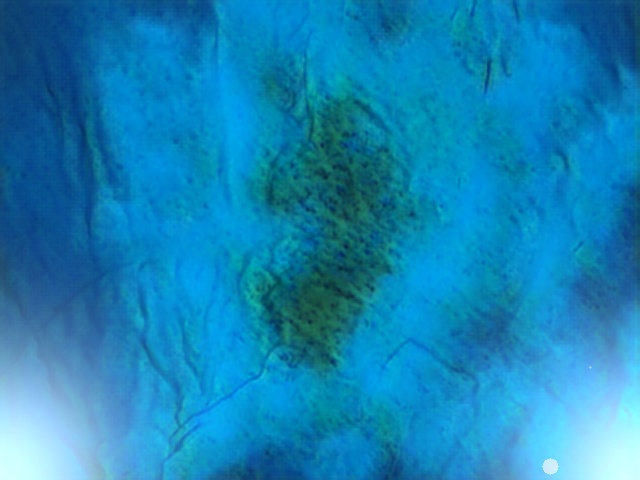}%
			\label{0007}}
		\hfil
	\end{center}
	\caption{Outputs of different underwater image simulators for the same scene.}{\tiny }
	\label{fig_evaluation}
\end{figure*}

We evaluate our deep sea image simulator by comparing with three state-of-the-art methods, which use in-air and depth images as the input to synthesize underwater images: UUV Simulator \cite{Manhaes_2016}, WaterGAN \cite{li2017watergan} and UW\_IMG\_SIM \cite{alvarez2019generation}. Due to the image size limitation from WaterGAN, all the evaluated images are simulated in the size of 640$\times$480, although our method does not have this limitation.

To render the realistic deep sea images close to the images shown in Fig. \ref{fig_lightCone}, we initialize the camera-light setups as:
two artificial spotlights which are 1m away from the camera on the left and right sides, both tilt 45$^{\circ}$ towards the image center. The real image was taken in the Niua region (Tonga) in south pacific ocean, according to the map of global distribution of Jerlov water types from \cite{johnson2012underwater}, water in this region belongs to type IB and the corresponding attenuation parameters are (0.37, 0.044, 0.035)[m$^{-1}$] for RGB channels. 
The simulation comparisons are given in Fig. \ref{fig_evaluation}. We create an in-air virtual scene with a sand texture, and simulate the corresponding underwater images by using the different methods. Since only our method considers the impact of lighting geometry configuration, the other methods are not able to add the shading effect on the texture image. To fairly compare our approach to others, we first add the in-air shading in the texture image and feed it to the other simulators, even though this in-air shading with a specific light RID is not available in any standard renderers. 

As it is shown in Fig. \ref{fig_evaluation}, the UUV Simulator is only able to render the attenuation effect based on the fog model without considering the impact of the light sources, the backscatter pattern caused by lighting is completely missing in their image. Their attenuation effect only considers the path from the scene points to the camera, which makes the rendered color also not conform to the deep sea scenario.
The same problem also occurs in the WaterGAN results, due to the lack of deep sea images with depth maps and ground truth in-air images, the GAN is trained using the parameters given in the official repository\footnote{\url{https://github.com/kskin/WaterGAN}} on the \emph{Port Royal, Jamaica} underwater dataset\footnote{\url{https://github.com/kskin/data}}. 
Therefore the color and the backscatter pattern of the light source is highly correlated with the training data which does not fulfill the setup in this evaluation case.
UW\_IMG\_SIM presents the backscatter pattern of the light source. However this effect is just adding the bright spots into the image without any physical interpretation, their direct signal component also has no dependence to the light source, which also is not realistic.
Our proposed approach captures all discussed effects present in real images better than the other methods, it not only renders the color much closer to the real image, but also simulates attenuated shading on the topography and back scatter caused by the artificial light sources which is missing in other approaches.

\section{Conclusion}
\label{Conclusions}
This paper presents a deep sea image simulation framework readily usable in current robotic simulation frameworks. 
It considers the effects caused by artificial spotlights, and provides good rendering results in deep sea scenarios at interactive framerates.
Earlier underwater imaging simulation solutions are either not physically accurate, or far from real-time to be integrated into a robotic simulation platform.
By detailed analysis of the deep sea image formation components, based on the Jaffe-McGlamery model, we propose several optimization strategies which enable us to achieve interactive performance and makes our deep sea imaging simulator fit to be integrated into the UUV simulator for prototyping or task planning.
This renderer has been applied for AUV lighting optimization in our later work \cite{song2021led}.
We release the source code of the deep sea image converter to the public to facilitate generation of training datasets and evaluation of underwater computer vision algorithms.

\subsubsection*{Acknowledgements.}
This publication has been funded by the German Research Foundation (Deutsche Forschungsgemeinschaft, DFG) Projektnummer 396311425, through the Emmy Noether Programme.
%
%
%
 \bibliographystyle{splncs04}
 \bibliography{references.bib}
%
%
%
%
%
\end{document}